\documentclass[preprint,3p,times,twocolumn]{elsarticle}

\usepackage{amsmath,amsfonts}
\usepackage{algorithm}
\usepackage{array}
\usepackage{textcomp}
\usepackage{stfloats}
\usepackage{subfigure}
\usepackage{url}
\usepackage{verbatim}
\usepackage{graphicx}
\usepackage{color}
\usepackage{soul, color, xcolor}
\usepackage{multirow}
\usepackage[colorlinks,linkcolor=red]{hyperref}
\usepackage{caption}
\sethlcolor{yellow}

\usepackage{algorithm}
\usepackage{algpseudocode}
\usepackage{amsmath}

\usepackage{amssymb}
\journal{xxx}

\begin{document}

\begin{frontmatter}



\title{LGEST: Dynamic Spatial-Spectral Expert Routing for Hyperspectral Image Classification}

\author[a,b]{Jiawen Wen}
\author[b]{Suixuan Qiu}
\author[c]{Zihang Luo}
\author[c]{Xiaofei Yang\corref{cor1}}
\author[c]{Haotian Shi}

\cortext[cor1]{Corresponding author: Xiaofei Yang\\
\hspace*{1.5em}\parbox[t]{\dimexpr\linewidth-1em\relax}{E-mail addresses: jwen341@connect.hkust-gz.edu.cn (J. Wen), qiusuixuan@mail.bnu.edu.cn (S. Qiu), 1275174812@e.gzhu.edu.cn (Z. Luo), xiaofeiyang@gzhu.edu.cn (X. Yang), shihaotian@gzhu.edu.cn (H. Shi)}}

\affiliation[a]{organization={Information Hub, The Hong Kong University of Science and Technology (Guangzhou)},
            city={Guangzhou},
            postcode={511453},
            country={China}}

\affiliation[b]{organization={Faculty of Geographical Science, Beijing Normal University},
            city={Beijing},
            postcode={100875},
            country={China}}

\affiliation[c]{organization={School of Electronic and Communication Engineering, Guangzhou University},
            addressline={Guangzhou University, Guangzhou},
            city={Guangzhou},
            postcode={510006},
            country={China}}

%% Abstract
\begin{abstract}
Deep learning methods, including Convolutional Neural Networks, Transformers and Mamba, have achieved remarkable success in hyperspectral image (HSI) classification. Nevertheless, existing methods exhibit inflexible integration of local-global representations, inadequate handling of spectral-spatial scale disparities across heterogeneous bands, and susceptibility to the Hughes phenomenon under high-dimensional sample heterogeneity. To address these challenges, we propose Local-Global Expert Spatial-Spectral Transformer (LGEST), a novel framework that synergistically combines three key innovations. The LGEST first employs a Deep Spatial-Spectral Autoencoder (DSAE) to generate compact yet discriminative embeddings through hierarchical nonlinear compression, preserving 3D neighborhood coherence while mitigating information loss in high-dimensional spaces. Secondly, a Cross-Interactive Mixed Expert Feature Pyramid (CIEM-FPN) leverages cross-attention mechanisms and residual mixture-of-experts layers to dynamically fuse multi-scale features, adaptively weighting spectral discriminability and spatial saliency through learnable gating functions. Finally, a Local-Global Expert System (LGES) processes decomposed features via sparsely activated expert pairs: convolutional sub-experts capture fine-grained textures, while transformer sub-experts model long-range contextual dependencies, with a routing controller dynamically selecting experts based on real-time feature saliency. Extensive experiments on four benchmark datasets demonstrate that LGEST consistently outperforms state-of-the-art methods.
\end{abstract}

%%Funding information


% \begin{highlights}
% \item Research highlight 1:
% First dynamic local-global expert routing framework for HSI classification, integrating sparsely activated convolutional and transformer sub-experts via confidence-aware gating to resolve spectral-spatial misalignment.

% \item Research highlight 2:
% Deep Spatial-Spectral Autoencoder (DSAE) with hierarchical nonlinear compression to generate compact 3D embeddings while preserving neighborhood coherence, mitigating the Hughes phenomenon.

% \item Research highlight 3:
% Cross-Interactive Mixed Expert Feature Pyramid (CIEM-FPN) with learnable cross-attention and residual mixture-of-experts layers, adaptively weighting multi-scale spectral and spatial cues for enhanced feature fusion.
% \end{highlights}

%% Keywords

\end{frontmatter}




\section{Introduction}
Hyperspectral images (HSIs) capture contiguous spectral bands across the electromagnetic spectrum, forming 3D data cubes rich in spatial and spectral information. These enable fine-grained material discrimination critical for military reconnaissance \cite{ardouin2007demonstration}, precision agriculture \cite{singh2020hyperspectral}, and mineral exploration \cite{peyghambari2021hyperspectral}. However, HSI classification faces inherent challenges: high dimensionality, spectral-spatial heterogeneity, and complex inter-band correlations that hinder accurate pixel-level categorization.
%% Use \subsubsection, \paragraph, \subparagraph commands to 
%% start 3rd, 4th and 5th level sections.
%% Refer following link for more details.
%% https://en.wikibooks.org/wiki/LaTeX/Document_Structure#Sectioning_commands
HSI classification aims to assign each pixel in the HSI to a specific land-cover class. The HSI classification methods could be divided into two categories: Traditional methods and Deep learning-based methods. Traditional methods rely on spectral signatures while ignoring spatial context. For example, Support Vector Machine (SVM)~\cite{li2011effective,chen2025ams} separate classes via optimal hyperplanes but falter with nonlinear data distributions. K-Nearest Neighbor (KNN) ~\cite{ma2010local} leverages spectral similarity metrics yet suffers from the ``curse of dimensionality". Multiple Linear Regression (MLR)~\cite{li2012semisupervised} class probabilities linearly but fails to capture spatial dependencies. However, these methods share critical limitations: inability to jointly model spectral-spatial features, sensitivity to noise, and poor scalability to high-dimensional data.

With the rapid development of deep learning, it has been widely applied to HSI classification, showing great potential in addressing the limitations of traditional methods. Convolutional neural networks (CNNs) are particularly effective in this task due to their hierarchical structure, which inherently captures local spatial patterns and spectral correlations. 2D-CNNs treat HSIs as two-dimensional images and mainly focus on spatial feature extraction. For example, Yang \textit{et al.}~\cite{yang2018hyperspectral} leveraging 2D-CNNs extract spatial features through stacked convolutional layers. In contrast, 3D-CNN frameworks address this issue by directly processing the 3D HSI data cube, jointly extracting spectral and spatial features. For example, Chen \textit{et al.}~\cite{chen2016deep} pioneered joint spectral-spatial feature learning, and Hamida \textit{et al.}~\cite{hamida20183} demonstrated enhanced discriminative power with deeper 3D architectures. Despite these advances, these approaches continue to face significant constraints. However, CNN-based methods also have problems. 2D-CNNs may not fully utilize the spectral information, and 3D-CNNs often suffer from high computational complexity and a large number of parameters, which can lead to overfitting, especially when the number of training samples is limited.

Transformer architectures have emerged as powerful solutions for HSI classification by leveraging self-attention mechanisms to model global dependencies. For example, Yang \textit{et al.}~\cite{yang2022spec} pioneers this approach by integrating 3D convolutions within transformer blocks to capture joint spatial-spectral relationships. However, it enforces uniform attention across spectral bands, failing to account for band-specific noise profiles. SpectralFormer~\cite{9627165} addresses spectral continuity through frequency-domain attention but neglects fine-grained spatial structures critical for boundary preservation. However, they also have limitations. Transformers typically require a large number of parameters and computational resources, which can be a burden for processing HSIs, especially on devices with limited computing power. They may also struggle with local feature extraction, as their self-attention mechanisms are more focused on global relationships.

Recently, Mamba-based methods have started to be explored in HSMambaI classification. Mamba, a new neural network architecture, has the potential to offer advantages in terms of computational efficiency and scalability. For example, MambaHSI~\cite{mambahsi} offer efficient long-range modeling but struggle with local texture preservation and spectral-spatial feature alignment. Additionally, in the context of image classification in general, Mixture-of-Experts (MoE) models have shown promise. MoE models consist of multiple "expert" sub-models, each specialized in different aspects of the data. In HSI classification, MoE can be used to handle the diverse characteristics of HSIs, such as different spectral patterns and spatial structures. For example, Switch Transformers~\cite{fedus2022switch} improve model capacity via sparse activation but remain unexplored for spectral-spatial adaptive routing in HSIs.

Although CNN-based, Transformer-based, and Mamba-based approaches have advanced hyperspectral image (HSI) classification, several challenges persist. CNN-based methods exhibit limitations: 2D CNNs insufficiently exploit spectral information, while 3D CNNs incur high computational complexity and a heightened risk of overfitting. Transformer-based approaches require considerable computational resources and tend to extract local features less effectively. Although Mamba-based methods demonstrate potential, they remain in the early exploratory phase and require further refinement to fully leverage the unique properties of HSIs. Furthermore, MoE-based approaches call for improved integration strategies to ensure cohesive collaboration among diverse experts and to effectively manage the complexity and high dimensionality of HSI data.

To address these challenges, we introduce LGEST, a novel method that comprehensively overcomes the limitations of existing approaches. Previous methods have either concentrated solely on local or global features or lacked an effective mechanism for integrating diverse types of information. In contrast, LGEST utilizes a local-global expert mechanism, wherein the local experts are designed to capture intricate spectral-spatial features, and the global experts employ a transformer-like architecture that excels at modeling long-range dependencies and capturing broad contextual information. By integrating both local and global experts, LGEST effectively harnesses the full spectrum of information present in hyperspectral images (HSIs). Moreover, inspired by the success of mixture-of-experts (MoE) frameworks in managing heterogeneous data characteristics, LGEST incorporates an MoE-inspired structure within its local and global components. This design enables individual experts to specialize in various aspects of HSI data, such as distinct spectral bands or spatial regions. For example, some local experts may focus on edge-related features within specific spectral ranges, while others capture texture variations across different spatial scales. Similarly, the global experts may specialize in modeling diverse global relationships, including overarching spectral trends or extensive spatial patterns. This innovative integration of local-global information, coupled with an MoE-inspired architecture, is anticipated to significantly enhance the classification performance of HSIs.

 In summary, the contributions of our proposed LGEST method are as follows:

\begin{enumerate}
  \item We propose LGEST, a novel hybrid framework for hyperspectral image classification that dynamically integrates local-global feature via sparsely activated experts, being the first to achieve such integration.
  
  \item We introduce a \textbf{Deep Spatial-Spectral Autoencoder (DSAE)} to learn compact and discriminative 3D spectral-spatial features through hierarchical nonlinear compression while preserving 3D neighborhood coherence.

  \item We develop a \textbf{Cross-Interactive Mixed Expert Feature Pyramid (CIEM-FPN)}, the novel MoE-enhanced feature pyramid, which uses cross-attention for adaptive spectral-spatial weighting to capture multi-scale features.

  \item We design a performance-oriented \textbf{Local-Global Expert System (LGES)} that resolves spectral-spatial misalignment via Confidence-aware routing to achieve 3D spectral-spatial features.
  
  \item We conduct extensive experiments on four hyperspectral datasets, demonstrating the superiority of LGEST over existing methods.
\end{enumerate}

\section{Related Work}\label{related}
\subsection{CNN-based Hyperspectral Image Classification Methods}
Convolutional neural networks (CNNs) have been widely applied to HSI classification due to their superior local perception capabilities and effectiveness in extracting spatial context information \cite{gao2019multi}. Many researchers have designed 2D CNNs for this task, such as the one proposed by Yang \textit{et al.}~\cite{yang2018hyperspectral} which consists of three 2D convolutional layers and a fully connected layer. However, these 2D CNNs extract features by separating spatial and spectral dimensions. To address this, Chen \textit{et al.} proposed a 3D CNN architecture using 3D convolution to extract spatial-spectral features \cite{chen2016deep}. Some other methods introduce 3D convolution to jointly process spectral and spatial information, such as the work by Hamida \textit{et al.} \cite{hamida20183} and the Synergistic Convolutional Neural Network (SyCNN) designed by Yang \textit{et al.} \cite{yang2020synergistic}.

While 3D CNNs jointly model spatial-spectral features, their fixed receptive fields struggle to balance local texture preservation (e.g., crop boundaries) and global context capture (e.g., large-scale mineral distributions). Hybrid designs like HybridSN~\cite{roy2019hybridsn} combine 3D and 2D convolutions but use rigid cascades, causing spectral over-smoothing in noisy bands. Recent work by Xi \textit{et al.}~\cite{xi2022dgssc} addresses class imbalance but retains static multi-scale fusion, limiting adaptability to heterogeneous scenes. These limitations underscore the need for dynamic architectures that co-optimize local and global feature hierarchies —a gap addressed by our LGES.

\subsection{Transformers-based Hyperspectral Image Classification Methods}
Recently, transformer networks have gained prominence in natural language processing and have since been applied to image classification tasks. Following the pioneering work of Dosovitskiy \textit{et al.}~\cite{dosovitskiy2020image}, several studies have addressed the issue of attention collapse in deep Vision Transformer (ViT) models, where feature maps tend to be identical in the top layers. To mitigate this, Zhou \textit{et al.}~\cite{zhou2021deepvit} proposed DeepViT, which introduced a re-attention mechanism using a learnable matrix to enhance attention map diversity. Yang \textit{et al.}~\cite{yang2023qtn} introduced a new Quaternion Transformer Network (QTN) designed to capture self-adaptive and long-range correlations in HSI classification. Conventional transformers often process 2D images as 1D sequences, limiting their ability to extract joint spatial-spectral features. To address this, researchers have incorporated convolution operations into transformer networks~\cite{huang2023ss,zhao2024hyper, 10122197, yang2022spec, sun2022spectral, song2024interactive, 10841983, zhang2023lightweight,11072185,feng2024s2eft}. For example, Yang \textit{et al.} introduced spectral-adaptive 3D convolutions within a transformer, creating the hyperspectral image transformer (HiT) network~\cite{yang2022spec} to capture local spatial-spectral features.

Vision transformers like HiT~\cite{yang2022spec} integrate 3D convolutions for local spectral-spatial modeling but enforce uniform attention across all bands, neglecting band-specific noise profiles. ISSFormer~\cite{song2024interactive} improves feature interaction but uses fixed fusion rules, failing to dynamically prioritize spatial or spectral cues (e.g., suppressing noisy urban bands while enhancing mineral-sensitive ones). Recent lightweight transformers~\cite{zhang2023lightweight} optimize efficiency but sacrifice spectral discriminability through aggressive token pruning. These works highlight a critical need for dynamic resource allocation —a capability uniquely enabled by LGEST’s Confidence-aware expert routing.

\subsection{Mamba-based Hyperspectral Image Classification Methods}
Mamba-based models have emerged as efficient alternatives for sequence modeling, with applications in HSI classification\cite{he20243dss,10919126,10965756,sheng2024dualmamba,cao2025few}. For example, MambaHSI~\cite{mambahsi} employs the Mamba architecture to capture long-range spatial–spectral dependencies, while 3DSS-Mamba\cite{he20243dss} introduces spectral space tag generation (SSTG) with 3D selective scanning. MambaLG\cite{pan2024hyperspectral} adopts a dual-branch structure to extract local and global background information. These studies primarily focus on aligning raw HSI inputs with the Mamba architecture. However, constructing state space models (SSMs) in this manner may reduce both local spatial invariance and computational efficiency. To address this, ConvMamba\cite{zhang2025convmamba} integrates convolution for feature and domain fusion, IGroupSS-Mamba\cite{he2024igroupss} employs Interval Group Space–Spectral Blocks (IGSSB) for contextual representation, and GraphMamba\cite{yang2024graphmamba} incorporates HyperMamba and SpatialGCN modules for parallel and adaptive processing. Despite their advantages in handling long-range contexts, Mamba-based methods face challenges in preserving local texture details and aligning spectral-spatial features, which are critical for accurate pixel-level classification in complex HSI scenes. This highlights the need for hybrid approaches that combine the efficiency of Mamba with mechanisms to enhance local feature preservation and multi-modal alignment.

\section{Proposed Approach}\label{method}
\subsection{Overview of LGEST}
Hyperspectral image data is a three-dimensional data structure that contains both spatial and spectral information. This information can be represented by a three-dimensional tensor $H\in {{\mathbb{R}}^{W\times H\times C}}$, where $W$ and $H$ represent spatial information and carry the HSI dataset's spatial dimensions, while $C$ represents spectral information. Numerous studies have demonstrated that integrating spatial and spectral information can improve HSI classification performance. Therefore, our model incorporates both local-global and spatial-spectral information to enhance hyperspectral image classification. We achieve this through the design of a cross-interactive mixed expert feature pyramid, which explores long-term spatial-spectral relationships between different image elements and the global spectral properties of individual image elements.

\begin{figure*}[htp] % 双栏图片
  \centering % 居中
  \includegraphics[width=0.9\textwidth]{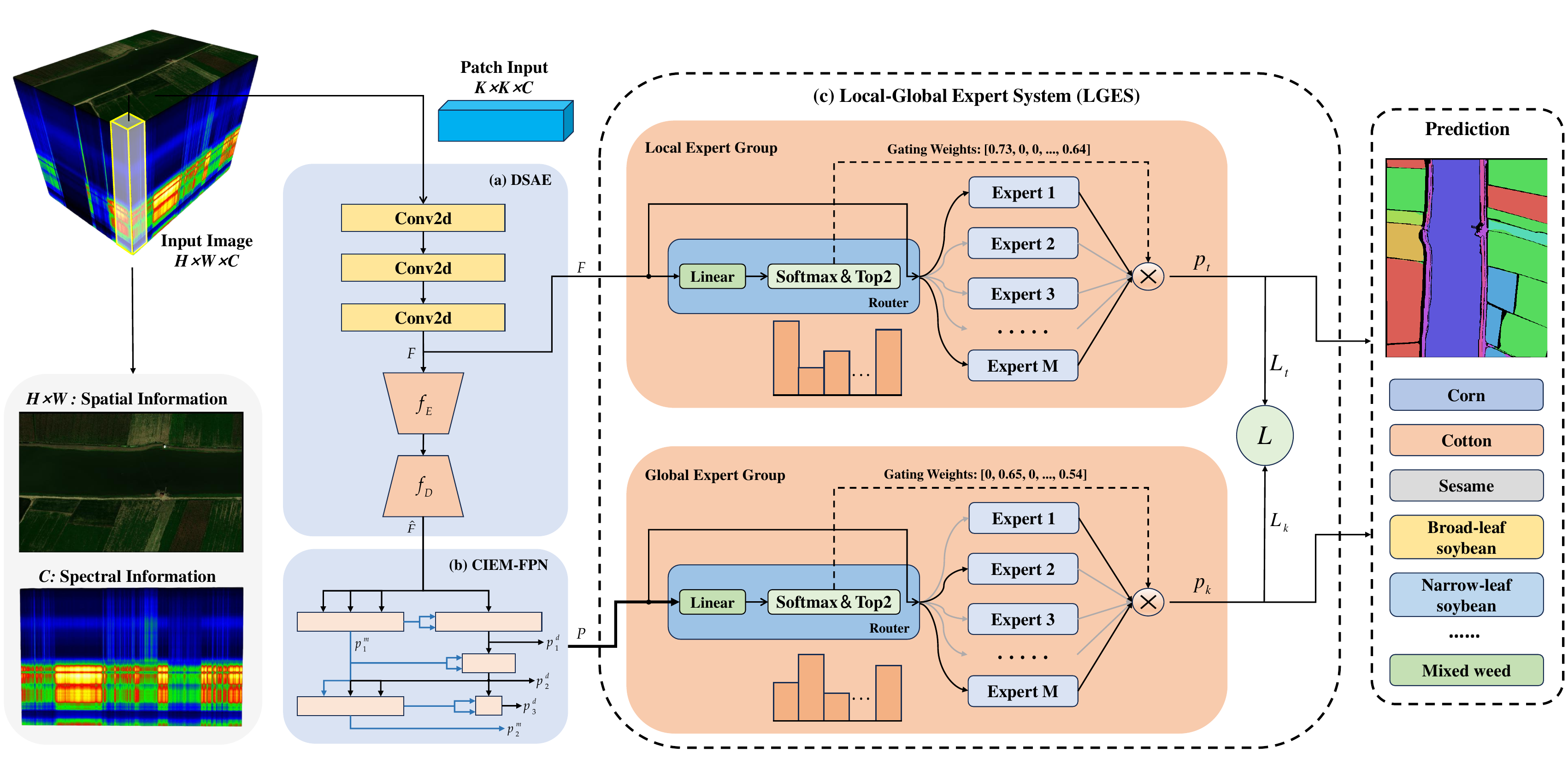}
  \caption{The architecture of proposed LGEST. The network processes input data with a deep spatial-spectral auto-encoder (DSAE), which serves as a local feature extractor. It then uses a cross-interactive mixed expert feature pyramid to fuse features across different scales via parallel and downsampling branches, enhancing interactions through the Cross-Interactive mixed Expert Feature Pyramid (CIEM-FPN). This is followed by the local-global expert system (LGES), divided into two expert groups handling local and global information. The router selects the activated expert from sub-experts within each group, which are linear layers.} 
  \label{fig_lgest}
\end{figure*}

As shown in Fig.~\ref{fig_lgest} and Algorithm~\ref{alg:lgest}, our model consists of three main components: the deep spatial-spectral auto-encoder (DSAE), the cross-interactive mixed expert feature pyramid (CIEM-FPN), and the local-global expert system (LGES). Multiple mixture of experts layers are incorporated into CIEM-FPN and the LGES. Critically, we employ a hierarchical decoupling strategy where these two expert systems address distinct challenges: physical heterogeneity in the feature space and semantic ambiguity in the decision space.

\begin{algorithm}[t!] 
\small % 缩小字体，让排版紧凑美观
\caption{Forward Pass of LGEST (Dynamic Spatial-Spectral Expert Routing)}
\label{alg:lgest}
\begin{algorithmic}[1] % [1] 表示每行都显示行号
\Require Hyperspectral image patch $X \in \mathbb{R}^{W \times H \times C}$, Ground truth label $y$
\Ensure Local predictions $P_t$, Global predictions $P_k$, Total Loss $L$

\Statex \textbf{// Phase 1: Local Feature Extraction via Deep Spatial-Spectral Auto-Encoder (DSAE)}
\State $F \leftarrow f_{conv}(X)$ \Comment{Extract local spatial features via Conv2d}
\State $K \leftarrow f_E(F)$ \Comment{Encoder: hierarchical nonlinear compression}
\State $\hat{F} \leftarrow f_D(K)$ \Comment{Decoder: reconstruct compact features}

\Statex \textbf{// Phase 2: Global Context Modeling via Cross-Interactive Mixed Expert Feature Pyramid (CIEM-FPN)}
\State Initialize feature lists for multi-scale fusion
\For{level $i = 1$ to $num\_fpn\_level$}
    \State $F_c \leftarrow CA(Q_i, K_i, V_i)$ \Comment{Dynamically adjust fusion weights}
    \State $P_i \leftarrow F_{RMoE}(F_c + Q_i)$ \Comment{Mitigate physical heterogeneity}
\EndFor
\State $P \leftarrow Concat(P_1, P_2, \dots, P_{num\_fpn\_level})$ \Comment{Aggregate multi-scale features}

\Statex \textbf{// Phase 3: Semantic Decision via Local-Global Expert System (LGES)}
\Statex \textit{// 3.1 Local Expert Group (Addressing local texture confusion)}
\State $S_{local} \leftarrow \text{Softmax}(W_{g\_local}^T F + b_{g\_local})$
\State $\mathcal{T}_{local} \leftarrow \text{Top2}(S_{local})$ \Comment{Select indices of the top-2 experts}
\State $S'_{local} \leftarrow S_{local} \cdot \text{Mask}(\mathcal{T}_{local})$ \Comment{Sparse activation masking}
\State $P_t \leftarrow \sum_{j \in \mathcal{T}_{local}} S'_{local}[j] \cdot Expert_j^{local}(F)$

\Statex \textit{// 3.2 Global Expert Group (Addressing global spectral variability)}
\State $S_{global} \leftarrow \text{Softmax}(W_{g\_global}^T P + b_{g\_global})$
\State $\mathcal{T}_{global} \leftarrow \text{Top2}(S_{global})$ \Comment{Select indices of the top-2 experts}
\State $S'_{global} \leftarrow S_{global} \cdot \text{Mask}(\mathcal{T}_{global})$ \Comment{Sparse activation masking}
\State $P_k \leftarrow \sum_{j \in \mathcal{T}_{global}} S'_{global}[j] \cdot Expert_j^{global}(P)$

\Statex \textbf{// Phase 4: Optimization}
\State $Loss(P_t) \leftarrow \text{CrossEntropy}(P_t, y)$
\State $Loss(P_k) \leftarrow \text{CrossEntropy}(P_k, y)$
\State $L \leftarrow \lambda \cdot Loss(P_t) + \beta \cdot Loss(P_k)$ \Comment{Compute final weighted loss}

\State \Return $P_t, P_k, L$
\end{algorithmic}
\end{algorithm}

Given an input $\{X,y\}$, a hyperspectral image paired with category-labeled data, we first pass $X$ through a DSAE to extract richer local spatial information from the HSI data and obtain more compact and representative features from the high-dimensional input. The encoder generates local features $F$ via three convolutions, which are then fed into our encoder-decoder structure to produce $\hat{F}$. $\hat{F}$ is then fed into the CIEM-FPN, which has a parallel and downsampling branch to extract features from the HSI data. Functioning in the feature space, CIEM-FPN employs "Feature Alignment Experts" within the cross-interactive mixed expert module (CIEM). These experts focus on mitigating physical heterogeneity by aligning feature distributions across different scales and modalities (Spectral vs. Spatial), thereby resolving feature misalignment. The CIEM enables softer interactions between diverse features, producing the output feature $P$. 

$F$ and $P$ represent local and global information. Finally, the Local-Global Expert System (LGES) transitions the processing framework into the semantic decision space. Diverging from the feature alignment objective of the preceding stage, LGES deploys 'Semantic Decision Experts' to dynamically modulate sub-expert activation. This mechanism is specifically engineered to mitigate semantic ambiguity (e.g., inter-class confusion) by adaptively arbitrating the distinct contributions of high-frequency details (e.g., fine-grained textures) and low-frequency contexts (e.g., broad spatial coverage) toward the final classification inference. The outputs ${{P}_{t}}$ and ${{P}_{k}}$ of the LGES are employed to compute the loss and predict the class probabilities based on the maximum score. The training loss is a weighted combination of the cross-entropy losses for both.
  \begin{align}
  Loss(\hat{x},y) &= -\sum\limits_{i=1}^{n}{y\log (\hat{x})} \label{eqn-1} \\
  L &= \lambda Loss({{P}_{t}})+\beta Loss({{P}_{k}}). \label{eqn-2}
  \end{align}
  Where $y$ represents the true label, while $\hat{x}$ signifies the probability distribution associated with the predicted class. The values of two hyperparameters, $\lambda$ and $\beta$, are utilized to maintain a balanced relationship between the different types of losses. Unless specified otherwise, the default settings for these hyperparameters are $\lambda = 1$ and $\beta = 0.5$.
	
\subsection{Deep Spatial-Spectral Auto-Encoder (DSAE)}
Hyperspectral images have rich spectral information, but their high dimensionality increases computational complexity and overfitting risk. As network layers grow, local spatial information (e.g., edges, texture) in HSI is lost, though it's crucial for pixel-by-pixel classification. To address these challenges, a dual-stream output architecture is employed in the DSAE module, where convolutional layers are first used to capture the raw spatial information and improve the local spatial dependency. The deep autoencoder compresses input data into a low-dimensional representation through successive nonlinear layers and then reconstructs it into the original high-dimensional space. This encoding–decoding process yields compact spatial–spectral features that retain essential information, thereby facilitating more effective extraction of global patterns.

% Our proposed DSAE module is shown in Fig.~\ref{fig_dsae}.

% 	\begin{figure}[htp] % htbp详见说明书，记得删除括号内容
% 		\centering % 居中
% 		\includegraphics[width=0.62\linewidth, height=0.3\textheight]{figures/DSAE.pdf}% 图片地址，可以pdf可以jpg，scale是缩放比例
% 		\caption{The Deep Spatial-Spectral Auto-Encoder (DSAE) module comprises a front convolutional layer, a convolutional encoder ${{f}_{E}}$, and a de-convolutional decoder ${{f}_{D}}$. The initial three convolutional layers are employed for feature compression and local feature extraction. The input size is that of a single patch from the Indian Pines dataset.} % 图片标题
% 		\label{fig_dsae} %
% 	\end{figure}

Specifically, the DSAE module consists of three convolutional layers, a convolutional encoder $f_E$, and a de-convolutional decoder $f_D$. Given an input feature $X$, the feature map $F$ is first generated by passing $X$ through the three convolutional layers. This output feature $F$ is then fed into the encoder $f_E$ and the decoder $f_D$. Both the encoder and decoder consist of a convolutional layer, a BatchNorm layer, and a LeakyReLU nonlinear activation layer.

The convolutional encoder progressively reduces the dimensionality of the feature map. Initially, the feature map $F$ is passed through the first convolutional layer, which generates a set of feature maps with $C_1$ output channels:
\begin{equation}
F_1 = \text{Conv}(X, C_1)
\end{equation}
Subsequent layers halve the number of output channels, progressively compressing the feature representation. After $L$ convolutional encoders, the output feature map $K$ has $\frac{C_0}{2^L}$ channels, where $C_0$ is the initial number of channels in $X$. The final output of the convolutional encoder is given by:
\begin{equation}
K = f_E(F_L) = \text{Conv}(F_L, C_L)
\end{equation}
where $C_L = \frac{C_0}{2^L}$ is the number of channels after $L$ layers.

The de-convolutional decoder reconstructs the original feature map by progressively increasing the dimensionality of $K$. Each layer in the decoder doubles the number of output channels compared to the previous layer. After $L$ de-convolutional decoders, the output feature map $\hat{F}$ retains the same dimensionality as $F$. The reconstruction process is as follows:
\begin{equation}
\hat{F} = f_D(K_L) = \text{DeConv}(K_L, C_L, C_0)
\end{equation}
where $C_0$ is the original number of channels in $F$ and $C_L = \frac{C_0}{2^L}$ is the number of channels at the $L$-th layer.

The overall DSAE module can be described by the following set of equations:
\begin{equation}
\begin{aligned}
  & F = F_{\text{conv}}(X), \\
  & K = f_E(F), \\
  & \hat{F} = f_D(K).
\end{aligned}
\end{equation}

Where $X$ is first transformed by the convolutional layers into $F$, then compressed into a low-dimensional feature $K$ by the encoder $f_E$, and finally reconstructed to $\hat{F}$ by the decoder $f_D$.

	\begin{figure*}[htp] % 双栏图片
		\centering % 居中
		\includegraphics[width=0.85\textwidth]{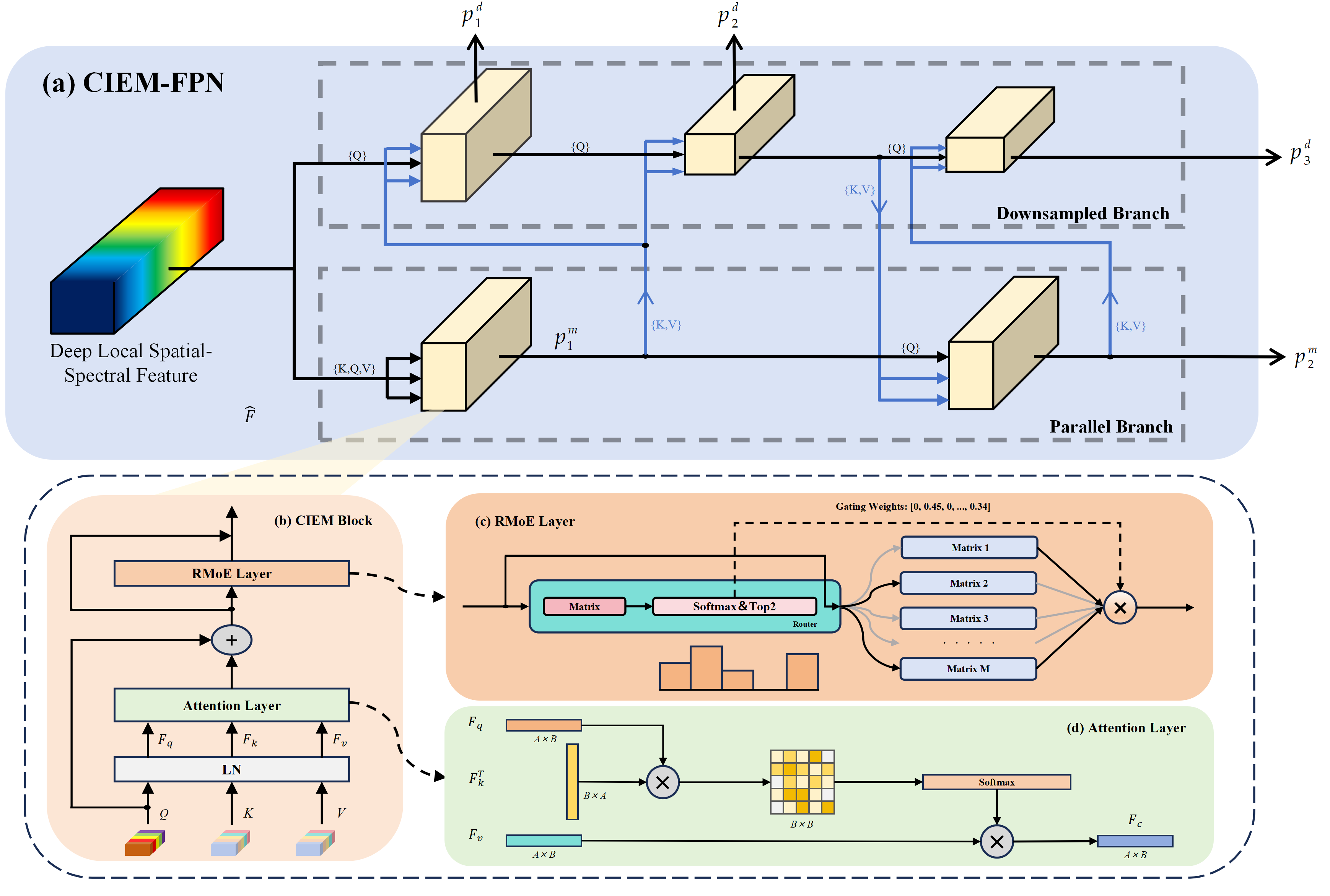}% 
		\caption{The structure of the cross-interactive mixed expert feature pyramid (CIEM-FPN). (a) shows the overall framework of CIEM-FPN, which incorporates a novel cross-interactive mixed expert module (CIEM) that enhances model comprehension and generation by creating associations between different input sequences.(b) presents the detailed structure of the CIEM module. To enable cross-interactive functionality, these inputs can originate from both the parallel structure and the upper output. (c) illustrates the RMoE layer, in which the experts are learnable matrices. (d) illustrates the cross-attention mechanism employed in CIEM.} % 图片标题
		\label{fig_fpn} %
	\end{figure*}

  %\cite{yang2023context,liu2023cb,li2023refine}
\subsection{Cross-Interactive Mixed Expert Feature Pyramid}
The FPN \cite{lin2017feature} has proven to be effective in both regression and prediction tasks in computer vision. However, its fusion process consists of basic up-sampling and down-sampling of features followed by addition, which often introduces noise and causes significant aliasing effects. Inspired by FPN, we introduce the CIEM-FPN, a parallel multiscale fusion module based on the cross-interactive mixed expert module (CIEM), to mitigate the noise generated by feature fusion at different scales and reveal deeper global relationships between spectral and spatial information across features. The CIEM-FPN fuses feature at different scales simultaneously through parallel and downsampling branches and enhances their interactions through a cross-attentional mechanism (Fig. \ref{fig_fpn}(a)). The cross-attention mechanism dynamically adjusts the fusion weights of features by calculating the correlation between the Query ($Q$), Key ($K$), and Value ($V$) components. This approach enables the model to emphasize relevant features while disregarding irrelevant or noisy information during the fusion process, resulting in finer feature fusion\cite{chen2021crossvit}. Compared to basic addition operations, the cross-attention mechanism offers a more flexible and precise method for feature interaction.

The CIEM-FPN comprises a basic unit, the CIEM module (Fig. \ref{fig_fpn} (b)), which includes Layer Normalization (LN) \cite{ba2016layer}, Cross-Attention \cite{hou2019cross}, and the RMoE layer. Fig. \ref{fig_fpn}(a) shows the $Q$, $K$, $V$ inputs of each CIEM module. For input features $Q$, $K$, and $V$, LN is first applied, followed by transformation into potential space expressions via multiplication with learnable parameter matrices, resulting in ${{F}_{q}}$, ${{F}_{k}}$, and ${{F}_{v}}$.
    %每个子行都有唯一编号
\begin{equation}\label{eqn-4}
  \left\{\begin{aligned}
    {{F}_{q}}&={{W}_{q}}(LN(Q)),\\
    {{F}_{k}}&={{W}_{k}}(LN(K)),\\
    {{F}_{v}}&={{W}_{v}}(LN(V)).
  \end{aligned}\right\}
\end{equation}
Where $Q$, $K$, and $V$ represent three inputs, while ${{W}_{q}}$, ${{W}_{k}}$, and ${{W}_{v}}$ represent three learnable parameter matrices. The generated ${{F}_{q}}$, ${{F}_{k}}$, and ${{F}_{v}}$ are produced by the cross-attention mechanism. As depicted in Fig. 3(d), the computational expression of this mechanism is given by:
\begin{equation}\label{eqn-5}
  CA(Q,K,V)=Softmax(\frac{{{F}_{q}}{{F}_{k}}^{T}}{\sqrt{C}}){{F}_{v}}.
\end{equation}
Where ${{F}_{q}}$, ${{F}_{k}}$, and ${{F}_{v}}$ are the input tokens, and $\frac{1}{\sqrt{c}}$ is the scaling factor. The dot product of ${{F}_{q}}$ and ${{F}_{k}}$ has an expectation of $0$ and a variance of $1$ when this scaling factor is used. Additionally, it prevents the softmax values from becoming overly large, thereby ensuring that the partial derivatives do not approach $0$. When $Q$, $K$, $V$ come from the same domain, cross-attention is self-attention.

We then add the generated cross-attention feature ${{F}_{c}}$ to the input $Q$ to create a more robust information representation. This representation is then fed into the proposed residual mixture of experts layer (RMoE Layer).

% \cite{shazeer2017outrageously}
The architecture of the RMoE Layer, as depicted in Fig. \ref{fig_fpn} (c), is composed of a residual branch \cite{fedus2022switch} and a main branch. The main branch involves the selection of two experts by a sparse gating function for computation, with their outputs modulated by the weights from the sparse gating function. The outputs of the two experts are combined to generate the main branch's output, which is then added to the residual branch to obtain the final output.

For an input $x$, the expression for the RMoE Layer is given below:
\begin{equation}\label{eqn-6}
    \left\{
    \begin{aligned}
      & G(x) = \text{Top2}(\text{Softmax}(x \cdot {W_g})), \\
      & {E_i}(x) = x \cdot {W_{e(i)}}, \\
      & y(x) = \sum_{i=1}^{n} G(x) E_i(x), \\
      & F_{\text{RMoE}}(x) = y(x) + x.
    \end{aligned}
    \right\}
\end{equation}
Where \( G(x) \) and \( E_i(x) \) represent the gating and expert functions, respectively, and \( W_g \) and \( W_{e(i)} \) are the learnable weight matrices. The primary role of the gating function is to compute the probability distribution across different experts and to select the two experts with the highest probabilities for activation. It is responsible for determining which tokens are to be allocated to which expert. The gating mechanism dynamically activates specific sub-experts based on the class of data input into the residual mixture of expert (RMoE) Layer, increasing the model's sparsity.

In the CIEM module, we replace the traditional transformer's FFN layer with the RMoE layer, which enables the model to select different experts (submodels) for processing various types of input. Consequently, the model can adaptively adjust its strategy for different types of input data and enhance its ability to process various spectral features. Through residual connections, the model retains the original information of the inputs when faced with unreliable expert outputs, thereby enhancing the robustness of the overall model. In summary, for the inputs  $Q$, $K$, and $V$, the expression for the CIEM module is given by:
    \begin{equation}\label{eqn-7}
      F_{\text{CIEM}}=F_{\text{RMoE}}\big(CA(F_{q}(Q), F_{k}(K), F_{v}(V))+Q\big)
      \end{equation}
Where $Q$, $K$, and $V$ denote the three inputs, ${{F}_{q}}$, ${{F}_{k}}$, and ${{F}_{v}}$ represent the potential space representations of the inputs, and ${{F}_{RMoE}}$ indicates the RMoE Layer. We construct the primary framework of the cross-interactive mixed expert feature pyramid (CIEM-FPN) based on the CIEM module. As shown in Fig.\ref{fig_fpn} (a), we denote $P_{i}^{m}$ as the output of the parallel branch (scale-invariant branch) and $P_{i}^{d}$ as the output of the downsampled branch, where $i$ represents the $i-th$ layer of the branches ($i \geq 1$). We use a step size of 2 for downsampling, and for CIEM-FPN inputs $\hat{F}\in {{\mathbb{R}}^{D\times C}}$, then $P_{1}^{m}\in {{\mathbb{R}}^{D\times C}}$ and $P_{1}^{d}\in {{\mathbb{R}}^{D\times \frac{C}{2}}}$. The complete CIEM-FPN expression is as follows:
\begin{equation}\label{eqn-8}
    \left\{
    \begin{aligned}
      & P_{i}^{m}={{F}_{CIEM}(\hat{F},\hat{F},\hat{F}),} & i=1 \\
      & P_{i}^{d}={{F}_{CIEM}(\hat{F},p_{1}^{m},P_{1}^{m}),} & i=1 \\
      & P_{i}^{m}={{F}_{CIEM}}(P_{i-1}^{m},P_{i-1}^{d},P_{i-1}^{d}), & i>1 \\
      & P_{i}^{d}={{F}_{CIEM}}(P_{i-1}^{d},P_{i-1}^{m},P_{i-1}^{m}), & i>1. \\
    \end{aligned}
    \right\}
\end{equation}
Where $P_{i}^{m}$ represents the output of the parallel branch (scale-invariant branch), $P_{i}^{d}$ represents the output of the downsampled branch, $i$ represents the $i-th$ layer of the branch ($i \geq 1$), and ${{F}_{CIEM}}$ is the CIEM module. Accordingly, we can simplify the expression of CIEM-FPN as:
\begin{equation}\label{eqn-9}
    \begin{aligned}
      {F}_{CIEM-FPN}(\hat{F}) & = \{p_{L/2}^{m}, p_{1}^{d}, p_{2}^{d}, p_{3}^{d}, \dots, p_{L}^{d}\} \\
          &= \{{{p}_{0}}, {{p}_{1}}, {{p}_{2}}, {{p}_{3}}, \dots, {{p}_{L}}\}.
    \end{aligned}
\end{equation}
Where $L$ indicates that there are $L$ layers in the pyramid. To maintain a linear gradient of the output features throughout the pyramid, we use the output of the last layer of the parallel branch $P_{2i-1}^{m}$ as the input to the first layer of CIEM-FPN. We concatenate all the feature vectors to obtain the final output vector $P$.
\begin{equation}\label{eqn-10}
P=\text{Concat}({{F}_{CIEM-FPN}}(\hat{F})).
% P=\text{Concat}_{\text{spectral}}({{F}_{CIEM-FPN}}(\hat{F}))
\end{equation}
Where $\hat{F}$ is the feature input to CIEM-FPN and $P$ is the final output of CIEM-FPN, $P$ incorporates enhanced multiscale global features from each layer of the pyramid. 
% However, these fused features still contain complex inter-class overlaps. Therefore, a specialized decision module is required to translate these aligned features into accurate categorical predictions.

\subsection{Local-Global Expert System (LGES)}

Although CIEM-FPN successfully constructs a fused spatial-spectral representation by aligning heterogeneous features, mapping these high-dimensional embeddings directly to land-cover classes remains challenging due to the 'Same Spectrum Different Object' phenomenon. Distinct from the feature-level fusion in CIEM-FPN, the LGES is designed as a decision-level refinement module. While CIEM-FPN focuses on mitigating physical discrepancies (e.g., scale and noise), LGES focuses on resolving semantic ambiguities. By decoupling the classification head into local-specific and global-specific pathways, LGES ensures that the final inference is robust against both local texture confusion and global spectral variability. This hierarchical expert design allows the model to separate feature alignment from semantic decision-making.

The Local-Global Expert System (LGES) is employed to independently process the local information provided by DSAE and the global information extracted by CIEM-FPN. Specifically, the LGES consists of two expert groups: the local expert group and the global expert group, both of which share the same structure.

As shown in Fig.~\ref{fig_lgest} (c), the system consists of two expert groups processing the preliminary local feature \(F\) and the extended multiscale global interactive feature \(P\). The structure of the two expert groups closely resembles that of the RMoE layer in CIEM, with the following distinctions: 1) In LGES, both the expert and gating mechanisms consist directly of the linear layer, and the output of the expert group, $P \in {\mathbb{R}^{{N \times n_{class}}}}$, is used directly for classification. 2) The expert group in LGES omits the residual linkage and outputs the expert’s decision results directly.

The expert group is designed to dynamically aggregate outputs from multiple expert networks, allowing for adaptive computation based on the input. Let the input feature vector be \( \mathbf{x} \in \mathbb{R}^d \), where \( d \) is the feature dimension. The layer consists of \( M \) expert networks, each represented as a linear transformation:
\begin{equation}\label{eqn-11}
f_i(\mathbf{x}) = \mathbf{W}_i^{T} \mathbf{x} + \mathbf{b}_i, \quad i = 1, 2, \dots, M,
\end{equation}
where \( \mathbf{W}_i \in \mathbb{R}^{d \times d_{\text{out}}} \) and \( \mathbf{b}_i \in \mathbb{R}^{d_{\text{out}}} \), with \( d_{\text{out}} \) being the output dimension. A gating mechanism computes the softmax-normalized weights for the experts:
\begin{equation}\label{eqn-12}
  \mathbf{s} = \text{softmax}(\mathbf{W}_g^{T} \mathbf{x} + \mathbf{b}_g),
\end{equation}
where \( \mathbf{W}_g \in \mathbb{R}^{d \times M} \) and \( \mathbf{b}_g \in \mathbb{R}^M \). Instead of using fixed routing, we employ a content-adaptive gating mechanism. The Top-2 softmax selection naturally filters out low-confidence experts, allowing the model to focus on the most relevant feature subspaces. To reduce computational overhead, only the top-2 experts with the highest weights are selected. Let \( \mathcal{T}_2 \subset \{1, 2, \dots, M\} \) denote the indices of the top-2 experts. A mask \( \mathbf{m} \in \mathbb{R}^M \) is applied such that:
\begin{equation}\label{eqn-13}
      \mathbf{m}[i] =
\begin{cases}
1, & \text{if } i \in \mathcal{T}_2, \\
0, & \text{otherwise}.
\end{cases}
\end{equation}

The weights for the top-2 experts are then refined as \( \mathbf{s}' = \mathbf{s} \odot \mathbf{m} \), where \( \odot \) represents element-wise multiplication. The outputs of all experts are stacked into a tensor \( \mathbf{E} \in \mathbb{R}^{N \times M \times d_{\text{out}}} \), where \( N \) is the batch size, and the final output is computed as:
 \begin{equation}\label{eqn-14}
         \mathbf{m}[i] =
    \begin{cases}
    \mathbf{y} = \mathbf{s}'^\top \mathbf{E}.
    \end{cases}
    \end{equation}

If the number of experts is less than 2, defined (\( M = 0,1 \)), the layer defaults to a single linear transformation \( f_0(\mathbf{x}) \). Formally, the output of the expert group can be expressed as:
 \begin{equation}\label{eqn-15}
    {f}_{expertgroup} =
    \begin{cases}
    f_0(\mathbf{x}), & \text{if } M = 0,1 \\
    \mathbf{s}'^\top \mathbf{E}, & \text{if } M > 1.
    \end{cases}
    \end{equation}
This design enables the expert group to adaptively utilize selected experts for efficient and flexible computation. The complete expression for the local-global expert system is as follows:
        \begin{equation}\label{eqn-16}
    \left\{
    \begin{aligned}
     & {{P}_{t}}={ {f}_{expertgroup}}(F), \\
 & {{P}_{k}}={ {f}_{expertgroup}}(P). \\
    \end{aligned}
    \right\}
    \end{equation}
Where \({f}_{expertgroup}\) denotes the expert group, \(F\) is the initial localized feature, and \(P\) is CIEM-FPN output. Each expert is trained on distinct data, allowing the model to integrate expert advantages for improved classification accuracy. This approach is particularly effective for high-dimensional, high-complexity hyperspectral data. Finally, \({P}_{t}\) and \({P}_{k}\) are directly employed for classification and loss calculation.

\section{Experiments}\label{experiment}
In this section, we evaluate the proposed LGEST on four publicly available HSI datasets: Indian Pines, Kennedy Space Center (KSC), Houston2013 and WHU-Hi-LongKou. To fully compare the effectiveness of our model, we compare LGEST with 13 other advanced HSI image classification methods. These include traditional HSI classification methods (e.g., SVM \cite{li2011effective}), CNNs-based HSI classification methods (e.g., 2D CNN \cite{yang2018hyperspectral}, 3D CNN \cite{hamida20183}, SYCNN \cite{yang2020synergistic}, HybridSN \cite{roy2019hybridsn}), and transformer-based HSI classification methods (e.g., ViT \cite{dosovitskiy2020image}, DeepViT \cite{zhou2021deepvit}, Cross ViT \cite{chen2021crossvit}, CaiT \cite{touvron2021going}, HiT \cite{yang2022spec}, SSFTT \cite{sun2022spectral}, Morphformer \cite{roy2023spectral}, SS-TMNet\cite{huang2023ss}, GSC-ViT \cite{zhao2024hyper}). The evaluation metrics are the Overall Accuracy (OA), Average Accuracy (AA), and Kappa coefficient ($\kappa$).

% The remainder of this section is organized as follows: First, we will introduce the adopted dataset and experimental settings in Section \ref{sec4.1}. In Section \ref{sec4.2}, we will show the results and analysis of our comparative experiments. In Section \ref{sec4.3}, we will analyze the complexity of our model. Lastly, in Section \ref{sec4.4}, we will design and analyze a series of ablation experiments around LGEST.

\subsection{Datasets and experimental settings}\label{sec4.1}
To evaluate the performance of LGEST, we selected four representative publicly available benchmark datasets: Indian Pines, Kennedy Space Center, Houston2013 and WHU-Hi-LongKou. Their detailed descriptions are given below:

\textbf{Indian Pines Dataset:} The Indian Pines dataset, collected in 1992 by the AVIRIS sensor in northwestern Indiana, contains 145 × 145 pixels with 200 usable spectral bands and provides 10,249 labeled samples across 16 land-cover classes at 20 m resolution.

\textbf{Kennedy Space Center (KSC):} The KSC dataset, collected in 1996 by NASA’s AVIRIS sensor over the Kennedy Space Center, contains 512 × 614 pixels with 176 high-quality bands spanning visible to near-infrared wavelengths. It provides 13 land-cover classes, including wetlands, forests, and agriculture, at 18 m resolution.

% The KSC dataset, acquired by NASA in 1996 with the AVIRIS sensor, covers the Kennedy Space Center area in Florida. It has a spatial resolution of 512 × 614 pixels, with 224 spectral bands, 176 of which are high-quality after removing noisy bands. The dataset, spanning visible to near-infrared wavelengths, includes 13 labeled categories of land cover, such as wetlands, forests, and agriculture, with each band having an 18-meter resolution. %The categories and training/testing distribution of this dataset are shown in Table \ref{tab2}.
% \begin{table}[htp]
% \centering
% \caption{Number of training/testing pixels for the KSC dataset.}\label{tab2}
% \begin{tabular}{|c|c|c|c|}
% \hline
% No. & Class Name & Training & Testing \\ \hline
% 1 & Scrub & 76 & 685 \\
% 2 & Willow swamp & 24 & 219 \\
% 3 & Cabbage palm hammock & 26 & 230 \\
% 4 & Cabbage palm/oak hammock & 25 & 227 \\
% 5 & Slash pine & 16 & 145 \\
% 6 & Oak/broadleaf hammock & 23 & 206 \\
% 7 & Hardwood swamp & 11 & 94 \\
% 8 & Graminoid marsh & 43 & 388 \\
% 9 & Spartina marsh & 52 & 468 \\
% 10 & Cattail marsh & 40 & 364 \\
% 11 & Salt marsh & 42 & 377 \\
% 12 & Mud flats & 50 & 453 \\
% 13 & Water & 93 & 834 \\ \hline
% \multicolumn{2}{|c|}{Total} & 521 & 4690 \\ \hline
% \end{tabular}
% \end{table}

\textbf{Houston2013:} The Houston2013 dataset, acquired in 2012 by the ITRES CASI-1500 sensor in Houston, contains 349 × 1905 pixels with 144 bands from visible to near-infrared and 15 land-cover classes, including trees, soil, and roads. It is widely used as a benchmark for multi-source fusion, classification, and urban planning studies.

% The Houston2013 dataset, acquired in 2012 by the ITRES CASI-1500 sensor in Houston, Texas, has a resolution of 349 × 1905 pixels and 144 spectral bands from visible to near-infrared. It contains 15 labeled land-cover classes, such as trees, soil, and roads. Widely used in research for multi-source data fusion, feature classification, and urban planning, it serves as an important benchmark for evaluating hyperspectral imaging algorithms.
% The categories and training/testing distributions of this dataset are shown in Table \ref{tab3}.
% \begin{table}[htp]
% \centering
% \caption{Number of training/testing pixels for the Houston2013 dataset.}\label{tab3}
% \begin{tabular}{|c|c|c|c|}
% \hline
% No. & Class Name & Training & Testing \\ \hline
% 1 & Healthy Grass & 125 & 1126 \\
% 2 & Stressed Grass & 125 & 1129 \\
% 3 & Synthetic Grass & 70 & 627 \\
% 4 & Trees & 124 & 1120 \\
% 5 & Soil & 124 & 1118 \\
% 6 & Water & 33 & 292 \\
% 7 & Residential & 127 & 1141 \\
% 8 & Commercial & 124 & 1120 \\
% 9 & Road & 125 & 1127 \\
% 10 & Highway & 123 & 1104 \\
% 11 & Railway & 124 & 1111 \\
% 12 & Parking Lot 1 & 123 & 1110 \\
% 13 & Parking Lot 2 & 47 & 422 \\
% 14 & Tennis Court & 43 & 385 \\
% 15 & Running Track & 63 & 594 \\ \hline
% \multicolumn{2}{|c|}{Total} & 1503 & 13526 \\ \hline
% \end{tabular}
% \end{table}

\textbf{WHU-Hi-LongKou:} The WHU-Hi-LongKou dataset, collected in 2018 in Longkou Town, China, using a UAV-mounted Headwall Nano-Hyperspec sensor at 500 m altitude, contains 550 × 400 pixels with 270 bands (400–1000 nm) at 0.463 m spatial resolution.

% The WHU-Hi-LongKou dataset, collected on July 17, 2018, in Longkou Town, Hubei Province, China, was acquired using a Headwall Nano-Hyperspec sensor mounted on a DJI Matrice 600 Pro UAV at a 500-meter altitude. The dataset has a resolution of 550 × 400 pixels, with 270 spectral bands ranging from 400 to 1000 nanometers and a spatial resolution of 0.463 meters.
% The categories and training/testing distributions of this dataset are shown in Table \ref{tabwh}.
% \begin{table}[htp]
% \centering
% \caption{Number of training/testing pixels for the WHU-Hi-LongKou dataset.}\label{tabwh}
% \begin{tabular}{|c|c|c|c|}
% \hline
% No. & Class Name & Training & Testing \\ \hline
% 1 & Corn & 35 & 34476 \\
% 2 & Cotton & 8 & 8366 \\
% 3 & Sesame & 3 & 3028 \\
% 4 & Broad-leaf soybean & 63 & 63149 \\
% 5 & Narrow-leaf soybean & 4 & 4147 \\
% 6 & Rice & 12 & 11842 \\
% 7 & Water & 67 & 66989 \\
% 8 & Roads and houses & 7 & 7117 \\
% 9 & Mixed weed & 5 & 5224 \\ \hline
% \multicolumn{2}{|c|}{Total} & 204 & 204338 \\ \hline
% \end{tabular}
% \end{table}

\textbf{Metrics:} To evaluate model performance, we use Overall Accuracy (OA), Average Accuracy (AA), and Kappa Coefficient ($\kappa$), common metrics in HSI image classification.

% OA is the ratio of correctly classified pixels to total pixels, reflecting overall classifier performance (values range from 0 to 1). AA averages the accuracy across categories, useful for imbalanced data. The Kappa Coefficient ($\kappa$) measures consistency by comparing actual and randomized results, with values from -1 to 1; higher values indicate better consistency. These metrics are defined using the number of categories ($m$), samples ($M$), and confusion matrix ($C$). The evaluation indicators are expressed as follows:
% \begin{align}
% OA &= \frac{1}{M}\sum\limits_{i=1}^{m}{{{C}_{ii}}}, \label{eqn-14} \\
% AA &= \frac{1}{m}\sum\limits_{i=1}^{m}{(\frac{{{C}_{ii}}}{\sum\limits_{j=1}^{m}{{{C}_{ij}}}})}, \label{eqn-15} \\
% Kappa(\kappa) &= \frac{M\sum\nolimits_{i=1}^{m}{{{C}_{ii}}-\sum\nolimits_{i=1}^{m}{({{C}_{i+}}\times {{C}_{+i}})}}}{{{M}^{2}}-\sum\nolimits_{i=1}^{m}{({{C}_{i+}}\times {{C}_{+i}})}}. \label{eqn-16}
% \end{align}
% Where ${{C}_{ii}}$ represents the number of samples on the diagonal of the confusion matrix i.e., the number of correctly classified samples; ${{C}_{i+}}$ denotes the total number of the $i-th$ row; and ${{C}_{+i}}$ represents the total number of the $i-th$ column.
  %\cite{kingma2014adam}
\textbf{Implementation details:} All models are implemented on Ubuntu 20.04.5 with an AMD EPYC 7542 CPU and an NVIDIA RTX 4090 GPU using PyTorch 1.12.1. Training uses the Adam optimizer with a learning rate of 1e-3, default settings, a batch size of 100, and 100 epochs. Each experiment is repeated 10 times.
To evaluate LGEST under limited labeled data, 10\% of samples are used for training on IP, KSC, and Houston2013, while only 0.1\% are used for the WHU-Hi-LongKou dataset due to its relatively simple spatial structure, highlighting robustness in extremely low-label settings.

% Our proposed LGEST and all other comparative models are implemented on an Ubuntu 20.04.5 LTS platform. Our CPU is an AMD EPYC 7542, and we use an NVIDIA RTX 4090 as our GPU. The deep learning framework utilized is PyTorch version 1.12.1. About hyperparameters, Adam is employed as the optimizer, with an initial learning rate of 1e-3. The remaining Adam hyperparameters are set to their default configuration. The batch size is set to 100, with 100 training rounds. It is noteworthy that all experiments were repeated 10 times.

\subsection{Analysis of Comparative Experiments with State-of-the-Art Models}\label{sec4.2}

% In this section, we present a comprehensive comparison of the proposed LGEST with other methods on four datasets.
% We perform a quantitative metrics analysis in Section \ref{sec4.2.1}, followed by an analysis of visualization results on the three datasets in Section \ref{sec4.2.2}. Finally, the t-SNE results of the different methods are analyzed in Section \ref{sec4.2.3}.

\subsubsection{Analysis of evaluation metrics results}\label{sec4.2.1}

This section will present a detailed analysis of the classification performance of the proposed LGEST in comparison to other HSI image classification methods. The results of the experiment are presented in Tables \ref{ID_ce}, \ref{KSC_ce}, \ref{HS_ce}, and \ref{WH_ce}, wherein the bolded evaluation metrics represent the optimal performance.

\begin{table*}[h]
    \centering
    \caption{Classification results for the IndianPines dataset (10\% training samples).}\label{ID_ce}
    \resizebox{\linewidth}{!}{
        \begin{tabular}{ccccccccccccccc}
        \hline
            \rule{0pt}{8.5pt}Class & SVM & 2D CNN & 3D CNN & HybridSN & SyCNN & ViT & Cross ViT & Deep ViT & CaiT & HiT & Morphformer & SS-TMNet & GSC-ViT & LGEST (ours) \\ \hline
        1 & 0.00 $\pm$ 0.00 & 15.56 $\pm$ 21.20 & 0.00 $\pm$ 0.00 & 0.00 $\pm$ 0.00 & 73.79 $\pm$ 8.25 & 0.00 $\pm$ 0.00 & 0.00 $\pm$ 0.00 & 7.95 $\pm$ 16.48 & 2.73 $\pm$ 5.45 & 14.19 $\pm$ 16.39 & 42.62 $\pm$ 23.14 & 68.73 $\pm$ 11.55 & 63.30 $\pm$ 30.15 & 86.46 $\pm$ 9.72  \\
        2 & 36.47 $\pm$2.36 & 88.73 $\pm$ 1.52 & 64.04 $\pm$ 3.58 & 60.05 $\pm$ 21.33 & 84.13 $\pm$ 1.33 & 48.68 $\pm$ 5.89 & 60.89 $\pm$ 5.29 & 78.18 $\pm$ 4.71 & 75.02 $\pm$ 2.33 & 86.47 $\pm$ 5.69 & 90.83 $\pm$ 1.71 & 87.35 $\pm$ 4.47 & 92.47 $\pm$ 1.39 & 92.51 $\pm$ 0.87
        \\
        3 & 0.00 $\pm$ 0.00 & 87.70 $\pm$ 1.61 & 66.25 $\pm$ 6.79 & 45.77 $\pm$ 28.93 & 85.64 $\pm$ 3.64 & 10.76 $\pm$ 5.95 & 26.65 $\pm$ 20.21 & 71.36 $\pm$ 9.57 & 70.62 $\pm$ 3.68 & 87.36 $\pm$ 2.90 & 87.99 $\pm$ 1.82 & 87.70 $\pm$ 2.20 & 88.78 $\pm$ 0.38 & 91.25  $\pm$ 1.44  \\
        4 & 0.00 $\pm$ 0.00 & 91.35 $\pm$5.33 & 14.33 $\pm$ 16.14 & 24.53 $\pm$ 33.52 & 91.30 $\pm$ 2.80 & 36.88 $\pm$ 16.35 & 61.66 $\pm$ 13.63 & 87.78 $\pm$ 5.11 & 86.50 $\pm$ 2.86 & 86.87 $\pm$ 10.28 & 91.78 $\pm$ 5.64 & 89.03 $\pm$ 5.24 & 91.98 $\pm$ 4.12 & 94.24 $\pm$ 2.88  \\
        5 & 0.23 $\pm$ 0.23 & 92.15 $\pm$ 1.98 & 83.09 $\pm$ 10.96 & 38.52 $\pm$ 31.57 & 88.75 $\pm$ 2.27 & 13.84 $\pm$ 11.51 & 56.83 $\pm$ 6.31 & 57.34 $\pm$ 13.77 & 49.31 $\pm$ 15.23 & 82.94 $\pm$ 11.04 & 91.53 $\pm$ 2.18 & 87.31 $\pm$ 1.96 & 94.13 $\pm$ 0.70 & 93.43 $\pm$ 1.98  \\
        6 & 66.95 $\pm$ 0.55 & 93.22 $\pm$ 1.32 & 85.85 $\pm$ 4.15 & 62.89 $\pm$ 16.27 & 90.52 $\pm$ 1.96 & 58.41 $\pm$ 5.82 & 73.54 $\pm$ 8.45 & 82.43 $\pm$ 2.75 & 73.46 $\pm$ 1.89 & 90.63 $\pm$ 2.20 & 92.38 $\pm$ 2.04 & 92.49 $\pm$ 1.26 & 94.14 $\pm$ 0.49 & 93.76 $\pm$ 1.22  \\
        7 & 0.00 $\pm$ 0.00 & 0.00 $\pm$ 0.00 & 0.00 $\pm$ 0.00 & 0.00 $\pm$ 0.00 & 0.00 $\pm$ 0.00 & 0.00 $\pm$ 0.00 & 0.00 $\pm$0.00 & 0.00 $\pm$ 0.00 & 0.00 $\pm$ 0.00 & 0.00 $\pm$ 0.00 & 19.05 $\pm$ 27.81 & 0.00 $\pm$ 0.00 & 12.97 $\pm$ 25.95 & 35.85 $\pm$ 34.95  \\
        8 & 87.92 $\pm$ 1.23 & 99.88 $\pm$ 0.18 & 99.15 $\pm$ 0.65 & 80.74 $\pm$ 20.82 & 98.24 $\pm$ 1.09 & 94.21 $\pm$ 2.36 & 95.82 $\pm$ 2.33 & 98.64 $\pm$ 0.82 & 97.45 $\pm$ 2.00 & 97.18 $\pm$ 0.93 & 99.06 $\pm$ 1.15 & 98.74 $\pm$ 0.76 & 99.20 $\pm$ 1.04 & 99.75  $\pm$ 0.31  \\
        9 & 0.00 $\pm$ 0.00 & 0.00 $\pm$ 0.00 & 0.00 $\pm$ 0.00 & 0.00 $\pm$ 0.00 & 0.00 $\pm$ 0.00 & 0.00 $\pm$ 0.00 & 0.00 $\pm$ 0.00 & 0.00 $\pm$ 0.00 & 0.00 $\pm$ 0.00 & 0.00 $\pm$ 0.00 & 0.00 $\pm$ 0.00 & 0.00 $\pm$ 0.00 & 0.00 $\pm$ 0.00 & 3.78 $\pm$ 8.44  \\
        10 & 0.00 $\pm$ 0.00 & 81.88 $\pm$ 1.92 & 69.05 $\pm$ 4.09 & 43.36 $\pm$ 36.17 & 79.72 $\pm$ 2.67 & 6.32 $\pm$ 6.37 & 46.57 $\pm$ 12.51 & 65.91 $\pm$ 9.35 & 52.40 $\pm$ 5.01 & 83.13 $\pm$ 1.95 & 86.36 $\pm$ 3.02 & 81.95 $\pm$ 1.98 & 86.98 $\pm$ 0.84 & 87.97 $\pm$ 2.15  \\
        11 & 57.36 $\pm$ 0.32 & 95.13 $\pm$0.76 & 77.73 $\pm$ 2.98 & 79.81 $\pm$ 10.05 & 90.99 $\pm$ 1.75 & 57.90 $\pm$ 1.64 & 69.06 $\pm$ 4.13 & 81.24 $\pm$ 5.03 & 79.35 $\pm$ 3.63 & 92.09 $\pm$ 2.18 & 94.79 $\pm$ 1.25 & 93.12 $\pm$ 2.50 & 96.25 $\pm$ 0.52 & 96.05  $\pm$ 0.56  \\
        12 & 0.00 $\pm$0.00 & 85.03 $\pm$ 1.84 & 58.62 $\pm$ 5.55 & 43.26 $\pm$ 29.51 & 85.02 $\pm$ 3.37 & 8.90 $\pm$ 8.82 & 37.54 $\pm$ 9.46 & 68.99 $\pm$ 8.99 & 69.29 $\pm$ 5.64 & 82.69 $\pm$ 7.88 & 86.36 $\pm$ 3.29 & 82.35 $\pm$ 5.53 & 86.41 $\pm$ 4.92 & 90.22  $\pm$ 1.58 \\
        13 & 0.00 $\pm$ 0.00 & 94.53 $\pm$ 3.96 & 44.36 $\pm$ 27.71 & 38.22 $\pm$ 37.50 & 90.08 $\pm$ 2.82 & 38.77 $\pm$ 33.09 & 43.71 $\pm$ 29.41 & 87.16 $\pm$ 7.86 & 85.00 $\pm$ 7.88 & 94.93 $\pm$ 2.27 & 90.39 $\pm$ 5.25 & 96.16 $\pm$ 3.04 & 95.33 $\pm$ 1.57 & 95.70  $\pm$ 1.69  \\
        14 & 82.83 $\pm$ 0.30 & 97.71 $\pm$ 0.83 & 92.18 $\pm$ 2.36 & 82.45 $\pm$ 8.16 & 93.68 $\pm$ 2.15 & 81.65 $\pm$ 2.80 & 87.17 $\pm$ 2.65 & 91.47 $\pm$ 1.70 & 89.82 $\pm$ 2.02 & 94.57 $\pm$ 2.27 & 97.27 $\pm$ 0.90 & 95.34 $\pm$ 0.75 & 98.46 $\pm$ 0.18 & 98.30 $\pm$ 0.50 \\
        15 & 0.00 $\pm$ 0.00 & 91.98 $\pm$ 3.20 & 58.58 $\pm$ 21.34 & 36.13 $\pm$ 29.56 & 86.99 $\pm$ 7.18 & 10.56 $\pm$ 19.73 & 68.63 $\pm$ 12.40 & 89.07 $\pm$ 5.16 & 74.28 $\pm$ 6.11 & 90.27 $\pm$ 3.91 & 93.23 $\pm$ 1.62 & 93.61 $\pm$ 1.62 & 95.48 $\pm$ 1.43 & 95.13 $\pm$ 1.74  \\
        16 & 44.68 $\pm$ 25.03 & 25.39 $\pm$ 24.13 & 0.00 $\pm$ 0.00 & 0.00 $\pm$ 0.00 & 58.16 $\pm$ 22.31 & 0.00 $\pm$ 0.00 & 4.70 $\pm$ 10.44 & 32.34 $\pm$ 26.41 & 0.00 $\pm$ 0.00 & 22.73 $\pm$ 24.47 & 35.77 $\pm$ 32.18 & 11.01 $\pm$ 15.62 & 24.92 $\pm$ 30.52 & 57.39 $\pm$ 23.02  \\ \hline
        AA(\%)  & 28.03 $\pm$ 1.23 & 70.12 $\pm$ 1.74 & 48.94 $\pm$ 2.95 & 40.83 $\pm$ 15.94 & 73.13 $\pm$ 2.22 & 30.96 $\pm$ 3.01 & 44.38 $\pm$ 5.63 & 60.31 $\pm$ 2.71 & 55.53 $\pm$ 1.11 & 68.08 $\pm$ 3.27 & 73.37 $\pm$ 2.55 & 71.48 $\pm$ 1.86 & 75.34 $\pm$ 2.94 & \textbf{80.72 $\pm$ 3.00}   \\
        OA(\%)  & 50.81 $\pm$ 0.26 & 91.09 $\pm$ 0.50 & 74.38 $\pm$ 2.46 & 65.64 $\pm$ 15.21 & 88.14 $\pm$ 1.19 & 51.93 $\pm$ 1.53 & 64.88 $\pm$ 4.51 & 79.48 $\pm$ 3.48 & 75.76 $\pm$ 1.52 & 88.77 $\pm$ 2.01 & 91.71 $\pm$ 0.87 & 89.77 $\pm$ 1.96 & 93.03 $\pm$ 0.46 & \textbf{93.69 $\pm$ 0.44}   \\
        Kappa(\%) & 40.10 $\pm$ 0.38 & 89.80 $\pm$ 0.56 & 70.13 $\pm$ 2.99 & 59.60 $\pm$ 18.41 & 86.45 $\pm$ 1.36 & 42.30 $\pm$ 2.27 & 58.80 $\pm$ 5.56 & 76.25 $\pm$ 4.09 & 71.87 $\pm$ 1.83 & 87.13 $\pm$ 2.35 & 90.52 $\pm$ 0.99 & 88.28 $\pm$ 2.27 & 92.03 $\pm$ 0.53 & \textbf{92.80 $\pm$ 0.50}  \\ \hline
    \end{tabular}
    }
\end{table*}

\begin{table*}[h]
    \centering
    \caption{Classification results for the KSC dataset (10\% training samples).}\label{KSC_ce}
    \resizebox{\linewidth}{!}{
        \begin{tabular}{ccccccccccccccc}
        \hline
            \rule{0pt}{8.5pt}Class & 2D CNN & 3D CNN &  HybridSN &  SyCNN & ViT & Cross ViT & Deep ViT & CaiT & HiT & SSFTT & Morphformer & SS-TMNet & GSC-ViT & LGEST (ours) \\ \hline
        1 & 86.06 $\pm$ 1.52 & 91.22 $\pm$ 2.83 & 92.73 $\pm$ 0.94 & 91.75 $\pm$ 0.98 & 55.30 $\pm$ 4.30 & 76.22 $\pm$ 2.71 & 78.72 $\pm$ 0.85 & 70.51 $\pm$ 5.77 & 93.72 $\pm$ 1.25 & 90.53 $\pm$ 6.93 & 90.61 $\pm$ 4.34 & 89.99 $\pm$ 7.37 & 95.00 $\pm$ 0.62 & 91.72 $\pm$ 1.61 \\
        2 & 57.62 $\pm$ 9.53 & 65.44 $\pm$ 11.37 & 49.31 $\pm$ 16.05 & 63.21 $\pm$ 8.43 & 0.00 $\pm$ 0.00 & 0.82 $\pm$ 2.08 & 1.45 $\pm$ 2.57 & 0.80 $\pm$ 1.61 & 22.61 $\pm$ 14.75 & 73.37 $\pm$ 13.53 & 72.10 $\pm$ 11.72 & 73.62 $\pm$ 4.14 & 67.53 $\pm$ 13.93 & 77.77 $\pm$ 4.23 \\
        3 & 51.43 $\pm$ 6.90 & 67.50 $\pm$ 15.67 & 77.63 $\pm$ 10.51 & 81.42 $\pm$ 7.37 & 0.00 $\pm$ 0.00 & 0.00 $\pm$ 0.00 & 0.00 $\pm$ 0.00 & 0.00 $\pm$ 0.00 & 73.18 $\pm$ 10.19 & 67.84 $\pm$ 24.07 & 71.44 $\pm$ 24.15 & 75.99 $\pm$ 7.31 & 70.91 $\pm$ 9.22 & 75.47 $\pm$ 5.66 \\
        4 & 41.69 $\pm$ 9.14 & 50.13 $\pm$ 9.48 & 51.99 $\pm$ 20.42 & 63.04 $\pm$ 12.31 & 29.10 $\pm$ 21.65 & 38.95 $\pm$ 17.36 & 50.43 $\pm$ 3.34 & 25.23 $\pm$ 20.85 & 31.94 $\pm$ 18.55 & 77.06 $\pm$ 13.31 & 74.65 $\pm$ 13.11 & 66.28 $\pm$ 11.48 & 60.58 $\pm$ 3.08 & 77.25 $\pm$ 5.77 \\
        5 & 64.78 $\pm$ 12.26 & 60.13 $\pm$ 11.54 & 37.63 $\pm$ 16.43 & 53.64 $\pm$ 12.49 & 5.60 $\pm$ 8.74 & 44.29 $\pm$ 24.05 & 53.83 $\pm$ 23.55 & 14.35 $\pm$ 16.74 & 89.49 $\pm$ 5.52 & 72.93 $\pm$ 14.29 & 67.16 $\pm$ 23.72 & 82.07 $\pm$ 7.40 & 71.87 $\pm$ 8.73 & 95.79 $\pm$ 2.45 \\
        6 & 53.46 $\pm$ 10.36 & 81.64 $\pm$ 5.69 & 73.99 $\pm$ 6.46 & 75.47 $\pm$ 9.69 & 0.00 $\pm$ 0.00 & 0.00 $\pm$ 0.00 & 3.21 $\pm$ 7.54 & 0.00 $\pm$ 0.00 & 87.78 $\pm$ 6.72 & 72.03 $\pm$ 27.06 & 77.59 $\pm$ 15.60 & 70.48 $\pm$ 17.80 & 72.81 $\pm$ 8.24 & 90.03 $\pm$ 7.67 \\
        7 & 48.23 $\pm$ 29.09 & 97.87 $\pm$ 1.90 & 96.20 $\pm$ 3.22 & 96.17 $\pm$ 4.83 & 0.00 $\pm$ 0.00 & 0.00 $\pm$ 0.00 & 0.00 $\pm$ 0.00 & 0.00 $\pm$ 0.00 & 95.98 $\pm$ 6.48 & 55.20 $\pm$ 40.80 & 84.43 $\pm$ 21.33 & 73.48 $\pm$ 37.24 & 47.84 $\pm$ 32.15 & 98.48 $\pm$ 3.53 \\
        8 & 75.55 $\pm$ 2.21 & 85.21 $\pm$ 5.48 & 89.12 $\pm$ 3.93 & 83.91 $\pm$ 3.50 & 26.83 $\pm$ 8.37 & 72.92 $\pm$ 9.64 & 64.29 $\pm$ 8.39 & 65.48 $\pm$ 19.10 & 89.32 $\pm$ 3.81 & 85.22 $\pm$ 8.23 & 85.95 $\pm$ 7.26 & 86.57 $\pm$ 6.56 & 72.47 $\pm$ 5.70 & 81.92 $\pm$ 5.15 \\
        9 & 87.42 $\pm$ 1.55 & 94.01 $\pm$ 3.61 & 95.25 $\pm$ 3.21 & 92.92 $\pm$ 2.49 & 39.88 $\pm$ 9.02 & 66.11 $\pm$ 1.41 & 62.46 $\pm$ 1.45 & 61.28 $\pm$ 3.73 & 95.07 $\pm$ 4.08 & 95.28 $\pm$ 3.05 & 94.30 $\pm$ 4.44 & 94.07 $\pm$ 2.36 & 85.60 $\pm$ 2.80 & 93.12 $\pm$ 3.23 \\
        10 & 93.27 $\pm$ 3.50 & 88.87 $\pm$ 5.00 & 84.44 $\pm$ 3.88 & 69.70 $\pm$ 4.08 & 56.11 $\pm$ 10.20 & 71.40 $\pm$ 3.41 & 69.05 $\pm$ 4.67 & 63.63 $\pm$ 13.22 & 80.12 $\pm$ 4.96 & 98.35 $\pm$ 3.08 & 96.57 $\pm$ 2.73 & 95.75 $\pm$ 5.61 & 98.57 $\pm$ 1.33 & 99.15 $\pm$ 1.54 \\
        11 & 99.46 $\pm$ 0.54 & 96.92 $\pm$ 3.86 & 90.53 $\pm$ 9.54 & 88.99 $\pm$ 3.27 & 37.71 $\pm$ 18.47 & 69.44 $\pm$ 10.09 & 77.88 $\pm$ 8.11 & 40.20 $\pm$ 24.47 & 99.96 $\pm$ 0.09 & 98.95 $\pm$ 1.33 & 99.72 $\pm$ 0.51 & 99.68 $\pm$ 0.43 & 99.95 $\pm$ 0.11 & 100.00 $\pm$ 0.00 \\
        12 & 93.08 $\pm$ 2.47 & 94.40 $\pm$ 2.65 & 82.86 $\pm$ 6.76 & 74.62 $\pm$ 4.86 & 79.40 $\pm$ 11.34 & 87.55 $\pm$ 3.06 & 89.14 $\pm$ 1.88 & 85.59 $\pm$ 7.37 & 71.93 $\pm$ 7.60 & 98.36 $\pm$ 1.34 & 96.77 $\pm$ 3.76 & 97.23 $\pm$ 2.48 & 97.52 $\pm$ 1.94 & 99.29 $\pm$ 1.03 \\
        13 & 100.00 $\pm$ 0.00 & 100.00 $\pm$ 0.00 & 100.00 $\pm$ 0.00 & 99.78 $\pm$ 0.42 & 97.87 $\pm$ 3.46 & 100.00 $\pm$ 0.00 & 100.00 $\pm$ 0.00 & 98.92 $\pm$ 1.48 & 95.58 $\pm$ 3.19 & 99.96 $\pm$ 0.07 & 100.00 $\pm$ 0.00 & 100.00 $\pm$ 0.00 & 100.00 $\pm$ 0.00 & 100.00 $\pm$ 0.00 \\ \hline
        AA(\%) & 71.62 $\pm$ 3.21 & 81.80 $\pm$ 3.10 & 78.01 $\pm$ 3.45 & 78.35 $\pm$ 2.97 & 35.96 $\pm$ 3.98 & 51.55 $\pm$ 2.59 & 52.81 $\pm$ 1.72 & 44.72 $\pm$ 5.53 & 79.54 $\pm$ 3.87 & 82.26 $\pm$ 7.14 & 84.02 $\pm$ 8.31 & 84.04 $\pm$ 5.89 & 79.33 $\pm$ 3.65 & \textbf{90.32 $\pm$ 1.89} \\
        OA(\%)  & 83.26 $\pm$ 1.39 & 87.98 $\pm$ 2.26 & 86.11 $\pm$ 2.25 & 84.03 $\pm$ 1.78 & 55.23 $\pm$ 4.10 & 70.45 $\pm$ 1.89 & 71.16 $\pm$ 1.13 & 64.81 $\pm$ 5.81 & 85.91 $\pm$ 2.83 & 90.69 $\pm$ 3.95 & 90.83 $\pm$ 4.41 & 90.21 $\pm$ 2.94 & 87.91 $\pm$ 1.55 & \textbf{92.59 $\pm$ 1.44} \\
        Kappa(\%) & 81.25 $\pm$ 1.56 & 86.59 $\pm$ 2.53 & 84.48 $\pm$ 2.52 & 82.18 $\pm$ 2.01 & 48.67 $\pm$ 4.85 & 66.58 $\pm$ 2.15 & 67.40 $\pm$ 1.27 & 60.00 $\pm$ 6.77 & 84.20 $\pm$ 3.20 & 89.57 $\pm$ 4.47 & 89.73 $\pm$ 4.97 & 89.06 $\pm$ 3.30 & 86.51 $\pm$ 1.74 & \textbf{91.75 $\pm$ 1.60} \\ \hline
    \end{tabular}
    }
\end{table*}

\begin{table*}[ht]
    \centering
    \caption{Classification results for the Houston2013 dataset (10\% training samples).} \label{HS_ce}
    \resizebox{\linewidth}{!}{
        \begin{tabular}{ccccccccccccccc}
        \hline
            \rule{0pt}{8.5pt}Class & SVM & 3D CNN &  HybridSN &  SyCNN & ViT & Cross ViT & Deep ViT & CaiT & RVT & HiT & Morphformer & SS-TMNet & GSC-ViT & LGEST (ours) \\ \hline
        1 & 90.94 $\pm$ 0.21 & 86.75 $\pm$ 1.72 & 83.54 $\pm$ 5.73 & 89.17 $\pm$ 1.11 & 79.78 $\pm$ 1.34 & 80.08 $\pm$ 2.46 & 86.50 $\pm$ 2.91 & 83.41 $\pm$ 3.72 & 88.36 $\pm$ 3.04 & 89.26 $\pm$ 1.68 & 92.63 $\pm$ 1.51 & 78.82 $\pm$ 24.36 & 91.79 $\pm$ 0.78 & 93.69 $\pm$ 1.94  \\
        2 & 96.56 $\pm$ 0.41 & 84.53 $\pm$ 3.22 & 83.45 $\pm$ 6.53 & 88.74 $\pm$ 2.34 & 63.64 $\pm$ 5.01 & 67.29 $\pm$ 5.82 & 76.70 $\pm$ 4.00 & 66.97 $\pm$ 4.18 & 89.50 $\pm$ 3.17 & 85.94 $\pm$ 2.11 & 93.44 $\pm$ 1.22 & 87.15 $\pm$ 10.45 & 93.66 $\pm$ 0.74 & 94.77 $\pm$ 1.18  \\
        3 & 59.63 $\pm$ 32.19 & 97.33 $\pm$ 1.59 & 93.32 $\pm$ 5.38 & 96.47 $\pm$ 1.60 & 95.24 $\pm$ 0.94 & 97.08 $\pm$ 1.46 & 97.51 $\pm$ 1.87 & 97.05 $\pm$ 0.82 & 97.50 $\pm$ 0.73  & 98.10 $\pm$ 1.03 & 98.48 $\pm$ 0.68 & 97.34 $\pm$ 1.56 & 98.16 $\pm$ 0.91 & 98.34 $\pm$ 1.34  \\
        4 & 93.87 $\pm$ 0.32 & 84.62 $\pm$ 3.48 & 78.82 $\pm$ 5.64 & 85.06 $\pm$ 2.97 & 64.98 $\pm$ 6.30 & 65.14 $\pm$ 7.32 & 76.68 $\pm$ 6.51 & 66.24 $\pm$ 5.60 & 84.97 $\pm$ 3.55 & 79.25 $\pm$ 4.31 & 93.93 $\pm$ 1.45 & 88.98 $\pm$ 7.77 & 94.24 $\pm$ 0.91 & 91.71 $\pm$ 1.62  \\
        5 & 90.97 $\pm$ 0.22 & 92.38 $\pm$ 3.09 & 91.79 $\pm$ 6.27 & 97.17 $\pm$ 1.08 & 87.35 $\pm$ 2.10 & 97.19 $\pm$ 1.05 & 98.47 $\pm$ 1.03 & 95.78 $\pm$ 0.89 & 99.44 $\pm$ 0.48 & 98.67 $\pm$ 0.64 & 99.21 $\pm$ 0.50 & 94.14 $\pm$ 10.30 & 99.83 $\pm$ 0.19 & 99.84 $\pm$ 0.15  \\
        6 & 89.94 $\pm$ 0.21 & 85.74 $\pm$ 1.15 & 85.82 $\pm$ 4.24 & 84.19 $\pm$ 2.62 & 79.63 $\pm$ 4.89 & 85.25 $\pm$ 1.34 & 83.03 $\pm$ 4.31 & 81.36 $\pm$ 3.86 & 84.47 $\pm$ 2.96 & 85.50 $\pm$ 0.89 & 90.48 $\pm$ 3.77 & 89.13 $\pm$ 3.97 & 88.23 $\pm$ 1.97 & 93.59 $\pm$ 1.34  \\
        7 & 68.20 $\pm$ 3.92 & 82.74 $\pm$ 2.43 & 74.68 $\pm$ 16.68 & 84.58 $\pm$ 1.64 & 76.11 $\pm$ 1.75 & 78.35 $\pm$ 2.83 & 79.15 $\pm$ 2.98 & 77.26 $\pm$ 2.75 & 85.91 $\pm$ 4.76 & 81.33 $\pm$ 3.59 & 94.29 $\pm$ 1.11 & 90.38 $\pm$ 4.78 & 94.54 $\pm$ 0.58 & 92.90 $\pm$ 1.90  \\
        8 & 54.07 $\pm$ 2.30 & 78.83 $\pm$ 3.65 & 81.37 $\pm$ 8.49 & 84.39 $\pm$ 2.27 & 68.37 $\pm$ 2.10 & 78.67 $\pm$ 4.86 & 84.78 $\pm$ 3.43 & 77.97 $\pm$ 3.09 & 93.39 $\pm$ 2.78 & 90.58 $\pm$ 2.67 & 94.79 $\pm$ 0.94 & 88.29 $\pm$ 12.62 & 95.05 $\pm$ 0.53 & 97.70 $\pm$ 0.52  \\
        9 & 62.16 $\pm$ 1.86 & 79.24 $\pm$ 4.63 & 48.18 $\pm$ 27.17 & 82.50 $\pm$ 0.98 & 68.85 $\pm$ 4.06 & 73.85 $\pm$ 5.29 & 81.16 $\pm$ 2.71 & 75.09 $\pm$ 2.26 & 85.93 $\pm$ 2.39 & 82.96 $\pm$ 1.88 & 91.72 $\pm$ 2.23 & 85.00 $\pm$ 9.28 & 93.16 $\pm$ 1.74 & 91.57 $\pm$ 1.31  \\
        10 & 38.44 $\pm$ 18.62 & 82.00 $\pm$ 4.61 & 83.08 $\pm$ 15.13 & 93.28 $\pm$ 2.40 & 72.94 $\pm$ 4.65 & 90.75 $\pm$ 2.31 & 90.18 $\pm$ 3.03 & 88.38 $\pm$ 1.23 & 91.26 $\pm$ 2.36 & 94.87 $\pm$ 1.68 & 97.93 $\pm$ 1.16 & 91.75 $\pm$ 11.30 & 98.01 $\pm$ 0.62 & 97.72 $\pm$ 1.42  \\
        11 & 63.58 $\pm$ 4.37 & 77.36 $\pm$ 7.39 & 66.87 $\pm$ 20.65 & 79.45 $\pm$ 3.59 & 64.59 $\pm$ 4.72 & 82.29 $\pm$ 4.34 & 84.94 $\pm$ 5.98 & 82.05 $\pm$ 3.33 & 90.80 $\pm$ 2.20 & 95.52 $\pm$ 1.85 & 98.15 $\pm$ 1.12 & 92.39 $\pm$ 11.71 & 96.88 $\pm$ 0.79 & 98.36 $\pm$ 0.92  \\
        12 & 36.27 $\pm$ 3.16 & 86.48 $\pm$ 2.90 & 84.45 $\pm$ 9.81 & 91.82 $\pm$ 2.34 & 66.68 $\pm$ 3.74 & 84.05 $\pm$ 2.35 & 89.57 $\pm$ 2.86 & 82.57 $\pm$ 3.85 & 89.46 $\pm$ 4.54 & 93.74 $\pm$ 2.43 & 95.42 $\pm$ 1.87 & 90.35 $\pm$ 12.27 & 96.60 $\pm$ 1.07 & 97.53 $\pm$ 0.63  \\
        13 & 0.00 $\pm$ 0.00 & 79.60 $\pm$ 4.81 & 63.93 $\pm$ 21.87 & 77.29 $\pm$ 4.65 & 29.73 $\pm$ 15.80 & 77.70 $\pm$ 6.18 & 86.16 $\pm$ 4.13 & 63.08 $\pm$ 9.59 & 83.28 $\pm$ 5.56 & 91.75 $\pm$ 2.48 & 93.17 $\pm$ 2.28 & 90.19 $\pm$ 6.77 & 92.41 $\pm$ 1.64 & 95.97 $\pm$ 1.05  \\
        14 & 85.30 $\pm$ 0.93 & 93.90 $\pm$ 3.94 & 81.82 $\pm$ 12.17 & 96.81 $\pm$ 1.51 & 82.64 $\pm$ 6.50 & 93.82 $\pm$ 3.86 & 95.27 $\pm$ 5.00 & 91.77 $\pm$ 4.29 & 99.04 $\pm$ 0.92 & 98.60 $\pm$ 0.84 & 98.84 $\pm$ 1.30 & 98.89 $\pm$ 2.15 & 98.45 $\pm$ 1.17 & 99.97 $\pm$ 0.05  \\
        15 & 98.52 $\pm$ 0.21 & 91.62 $\pm$ 2.43 & 75.10 $\pm$ 18.21 & 90.98 $\pm$ 4.09 & 75.32 $\pm$ 4.62 & 93.41 $\pm$ 2.50 & 94.59 $\pm$ 4.05 & 90.32 $\pm$ 5.29 & 95.91 $\pm$ 1.08 & 96.96 $\pm$ 1.30 & 97.53 $\pm$ 0.69 & 96.35 $\pm$ 1.65 & 97.30 $\pm$ 1.32 & 97.66 $\pm$ 1.65  \\\hline
        AA(\%)  & 70.49 $\pm$ 2.61 & 85.13 $\pm$ 1.23 & 80.25 $\pm$ 7.75 & 88.15 $\pm$ 0.52 & 71.27 $\pm$ 2.08 & 82.52 $\pm$ 1.81 & 86.24 $\pm$ 1.46 & 80.20 $\pm$ 1.97 & 91.08 $\pm$ 0.85 & 90.39 $\pm$ 1.23 & 94.92 $\pm$ 0.61 & 90.17 $\pm$ 8.45 & 94.74 $\pm$ 0.40 & \textbf{95.88 $\pm$ 0.40}  \\
        OA(\%)  & 71.83 $\pm$ 2.16 & 84.78 $\pm$ 1.29 & 79.49 $\pm$ 7.43 & 87.95 $\pm$ 0.67 & 72.30 $\pm$ 1.83 & 81.59 $\pm$ 1.89 & 86.07 $\pm$ 1.63 & 80.81 $\pm$ 1.46 & 90.42 $\pm$ 1.07 & 90.14 $\pm$ 1.51 & 95.34 $\pm$ 0.57 & 90.00 $\pm$ 9.45 & 95.44 $\pm$ 0.16 & \textbf{95.86 $\pm$ 0.58}  \\
        Kappa(\%) & 69.47 $\pm$ 2.36 & 83.55 $\pm$ 1.39 & 77.85 $\pm$ 8.03 & 86.98 $\pm$ 0.72 & 70.01 $\pm$ 1.99 & 80.09 $\pm$ 2.05 & 84.93 $\pm$ 1.77 & 79.22 $\pm$ 1.59 & 89.65 $\pm$ 1.16 & 89.34 $\pm$ 1.63 & 94.96 $\pm$ 0.61 & 89.18 $\pm$ 10.22 & 95.07 $\pm$ 0.17 & \textbf{95.53 $\pm$ 0.63}  \\\hline
    \end{tabular}
    }
\end{table*}

\begin{table*}[h]
    \centering
    \caption{Classification results for the WHU-Hi-LongKou dataset (0.1\% training samples).}\label{WH_ce}
    \resizebox{\linewidth}{!}{
        \begin{tabular}{ccccccccccccccc}
        \hline
        \rule{0pt}{8.5pt}Class & SVM & 2D CNN & 3D CNN & HybridSN & SyCNN & ViT & Cross ViT & Deep ViT & SSFTT & HiT & Morphformer & SS-TMNet & GSC-ViT & LGEST (ours) \\ \hline
        1 & 74.34 $\pm$ 1.28 & 99.13 $\pm$ 0.31 & 95.12 $\pm$ 1.88 & 87.72 $\pm$ 22.49 & 97.26 $\pm$ 0.95 & 85.94 $\pm$ 3.12 & 89.96 $\pm$ 3.00 & 92.45 $\pm$ 2.62 & 98.31 $\pm$ 0.78 & 97.34 $\pm$ 2.65 & 98.36 $\pm$ 0.93 & 99.10 $\pm$ 0.51 & 99.20 $\pm$ 0.41 & 99.30 $\pm$ 0.27 \\
        2 & 27.18 $\pm$ 14.88 & 64.75 $\pm$ 7.57 & 59.79 $\pm$ 13.99 & 50.97 $\pm$ 28.43 & 83.74 $\pm$ 5.91 & 58.59 $\pm$ 3.23 & 64.29 $\pm$ 6.50 & 67.76 $\pm$ 7.26 & 75.38 $\pm$ 27.09 & 75.16 $\pm$ 9.31 & 87.14 $\pm$ 4.97 & 81.33 $\pm$ 3.20 & 84.17 $\pm$ 7.31 & 86.52 $\pm$ 3.67 \\
        3 & 0.00 $\pm$ 0.00 & 0.00 $\pm$ 0.00 & 0.00 $\pm$ 0.00 & 1.02 $\pm$ 2.82 & 3.11 $\pm$ 4.96 & 0.00 $\pm$ 0.00 & 0.00 $\pm$ 0.00 & 0.00 $\pm$ 0.00 & 2.39 $\pm$ 5.47 & 9.61 $\pm$ 17.94 & 26.74 $\pm$ 31.27 & 20.60 $\pm$ 31.25 & 48.96 $\pm$ 40.50 & 74.51 $\pm$ 24.81 \\
        4 & 85.37 $\pm$ 0.56 & 91.14 $\pm$ 2.13 & 88.82 $\pm$ 0.66 & 90.01 $\pm$ 5.91 & 93.53 $\pm$ 0.69 & 90.76 $\pm$ 0.48 & 92.13 $\pm$ 0.84 & 92.51 $\pm$ 0.91 & 94.44 $\pm$ 1.01 & 94.04 $\pm$ 1.77 & 94.45 $\pm$ 1.46 & 94.74 $\pm$ 1.13 & 96.88 $\pm$ 1.21 & 96.68 $\pm$ 0.98 \\
        5 & 0.00 $\pm$ 0.00 & 19.05 $\pm$ 21.20 & 44.35 $\pm$ 19.88 & 19.09 $\pm$ 27.35 & 57.19 $\pm$ 15.82 & 0.00 $\pm$ 0.01 & 7.92 $\pm$ 9.82 & 10.18 $\pm$ 8.02 & 37.26 $\pm$ 30.46 & 38.35 $\pm$ 14.23 & 57.82 $\pm$ 14.87 & 62.91 $\pm$ 18.67 & 60.47 $\pm$ 28.11 & 63.38 $\pm$ 16.32 \\
        6 & 1.86 $\pm$ 3.11 & 96.45 $\pm$ 1.71 & 92.14 $\pm$ 3.78 & 69.91 $\pm$ 35.94 & 92.14 $\pm$ 3.81 & 49.70 $\pm$ 28.56 & 71.19 $\pm$ 21.21 & 75.20 $\pm$ 18.12 & 96.02 $\pm$ 1.84 & 95.44 $\pm$ 2.54 & 96.77 $\pm$ 1.38 & 96.05 $\pm$ 1.96 & 95.46 $\pm$ 3.72 & 95.40 $\pm$ 1.78 \\
        7 & 99.64 $\pm$ 0.07 & 98.90 $\pm$ 0.53 & 99.08 $\pm$ 0.47 & 94.29 $\pm$ 2.36 & 98.77 $\pm$ 0.82 & 98.81 $\pm$ 0.35 & 99.32 $\pm$ 0.27 & 99.22 $\pm$ 0.25 & 97.58 $\pm$ 1.17 & 97.69 $\pm$ 0.64 & 98.73 $\pm$ 0.68 & 98.62 $\pm$ 0.70 & 99.58 $\pm$ 0.26 & 99.08 $\pm$ 0.50 \\
        8 & 53.22 $\pm$ 7.06 & 70.66 $\pm$ 7.33 & 49.20 $\pm$ 6.09 & 36.02 $\pm$ 21.45 & 62.39 $\pm$ 9.75 & 67.08 $\pm$ 3.58 & 67.19 $\pm$ 4.54 & 61.26 $\pm$ 10.73 & 57.93 $\pm$ 13.55 & 76.45 $\pm$ 3.72 & 66.03 $\pm$ 11.18 & 74.95 $\pm$ 4.38 & 70.39 $\pm$ 5.14 & 77.07 $\pm$ 4.79 \\
        9 & 0.00 $\pm$ 0.00 & 21.95 $\pm$ 19.93 & 28.23 $\pm$ 26.99 & 15.00 $\pm$ 19.24 & 58.03 $\pm$ 19.14 & 17.53 $\pm$ 16.93 & 47.72 $\pm$ 13.87 & 49.91 $\pm$ 14.56 & 23.32 $\pm$ 20.92 & 44.58 $\pm$ 13.62 & 45.97 $\pm$ 23.69 & 54.39 $\pm$ 15.41 & 63.83 $\pm$ 16.97 & 70.91 $\pm$ 6.76 \\ \hline
        AA(\%) & 40.14 $\pm$ 1.51 & 61.31 $\pm$ 3.86 & 58.46 $\pm$ 3.06 & 51.50 $\pm$ 12.80 & 70.04 $\pm$ 3.49 & 52.62 $\pm$ 3.09 & 59.46 $\pm$ 3.77 & 60.35 $\pm$ 3.58 & 64.30 $\pm$ 5.21 & 67.69 $\pm$ 4.28 & 71.81 $\pm$ 5.39 & 73.23 $\pm$ 4.31 & 78.90 $\pm$ 6.60 & \textbf{82.14 $\pm$ 3.19} \\
        OA(\%) & 79.98 $\pm$ 0.48 & 91.17 $\pm$ 1.11 & 89.40 $\pm$ 0.79 & 85.58 $\pm$ 7.17 & 92.40 $\pm$ 0.70 & 86.12 $\pm$ 1.43 & 88.61 $\pm$ 1.54 & 89.15 $\pm$ 1.31 & 91.92 $\pm$ 1.50 & 92.28 $\pm$ 1.66 & 93.54 $\pm$ 1.21 & 93.78 $\pm$ 0.76 & 95.01 $\pm$ 0.98 & \textbf{95.40 $\pm$ 0.50} \\
        Kappa(\%) & 72.76 $\pm$ 0.62 & 88.12 $\pm$ 1.57 & 85.64 $\pm$ 1.09 & 80.15 $\pm$ 10.49 & 89.88 $\pm$ 0.96 & 81.31 $\pm$ 1.98 & 84.76 $\pm$ 2.11 & 85.51 $\pm$ 1.80 & 89.16 $\pm$ 2.06 & 89.69 $\pm$ 2.24 & 91.37 $\pm$ 1.66 & 91.71 $\pm$ 1.05 & 93.40 $\pm$ 1.31 & \textbf{93.91 $\pm$ 0.66} \\ \hline
        \end{tabular}
    }
\end{table*}

LGEST achieved superior performance across four datasets. On IndianPines, it improved AA to 80.72 $\pm$ 3.00\% (+5.38\%) and uniquely classified class 9 with 12.22\% accuracy. On KSC, it reached 92.59 $\pm$ 1.44\% OA, outperforming convolutional methods by 7.98\%, suggesting their limited use of global information. On Houston2013, LGEST improved AA by 9.64\% to 95.88 $\pm$ 0.40\%, aided by pyramidal interactions and rich local features, with gains up to 12.91\%. On WHU-Hi-LongKou, although slightly lower in OA than the second-best model, it improved AA by 3.24\%. Overall, LGEST excelled by combining DSAE for local features, CIEM-FPN for global spatial–spectral interactions, and a local–global expert system for specialized processing.

\subsubsection{Analysis of visualization results}\label{sec4.2.2}
\begin{figure*}[!t] % 双栏图片 \ref{class_ksc},
  \centering % 居中
  \includegraphics[width=1\textwidth]{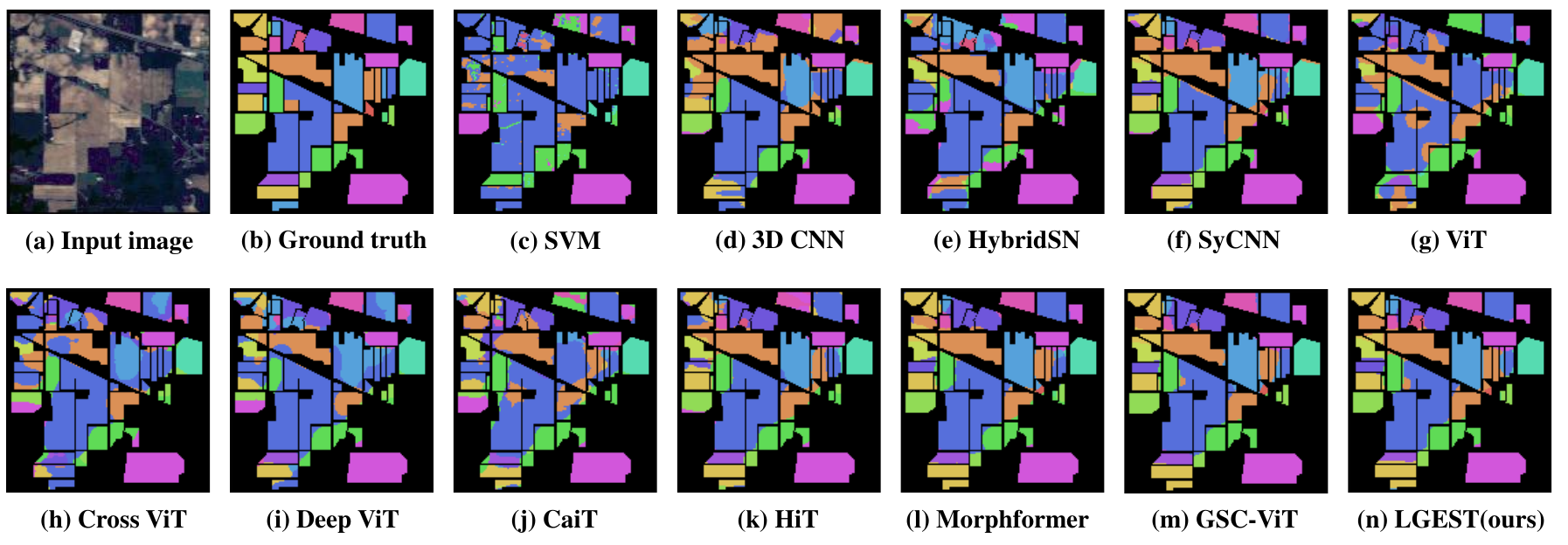}% 图片地址，可以pdf可以jpg，scale是缩放比例
  \caption{Classification map of different methods on the IndianPines dataset (with 10\% training samples).} % 图片标题
  \label{class_ip} %
\end{figure*}

\begin{figure*}[!t] % 双栏图片 \ref{class_ksc},
  \centering % 居中
  \includegraphics[width=1\textwidth]{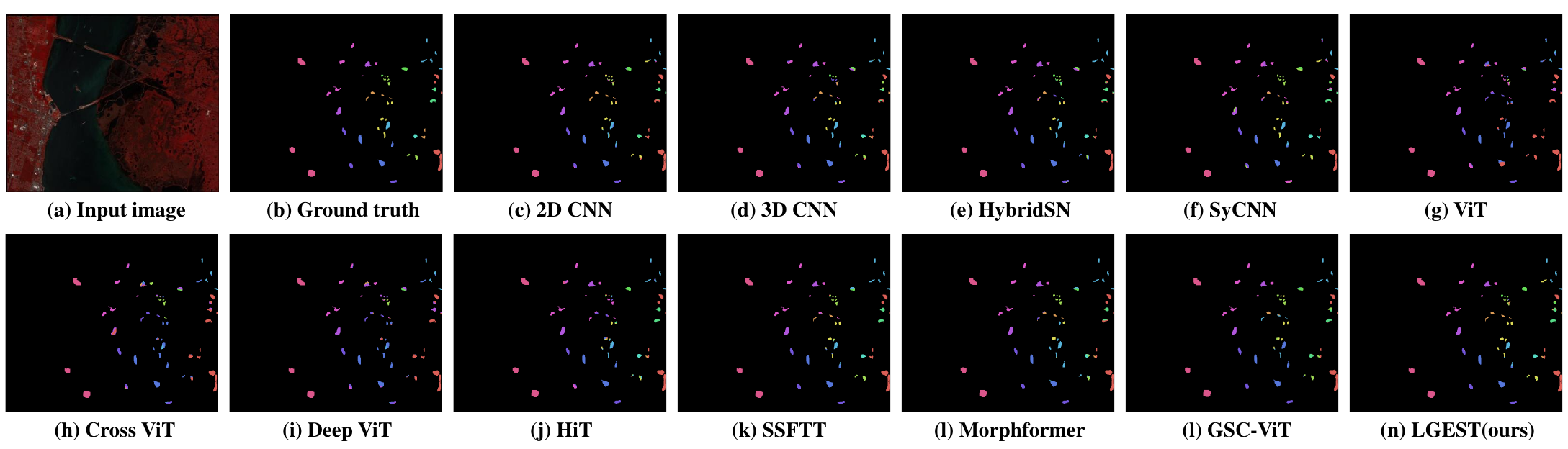}% 图片地址，可以pdf可以jpg，scale是缩放比例
  \caption{Classification map of different methods on the KSC dataset (with 10\% training samples).} % 图片标题
  \label{class_ksc} %
\end{figure*}

\begin{figure*}[!t] % 双栏图片
  \centering % 居中
  \includegraphics[width=1\textwidth,height=0.25\textheight]{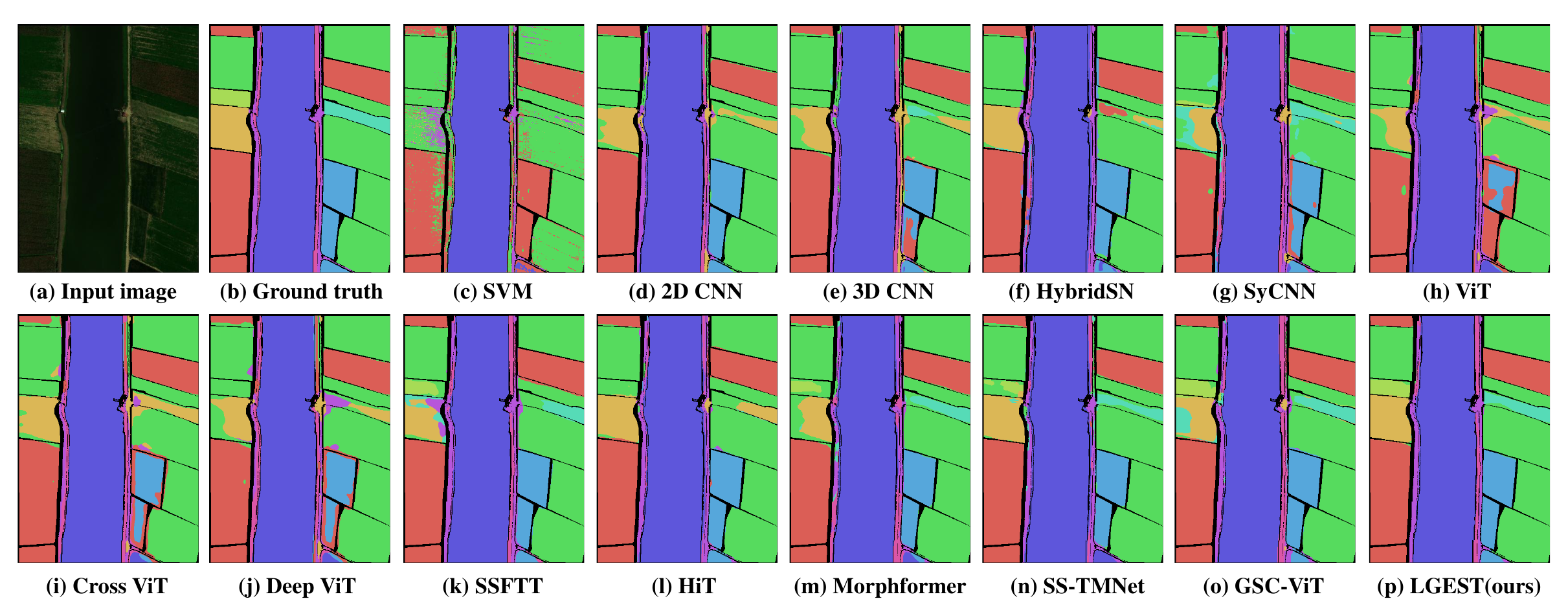}% 
  \caption{Classification map of different methods on the WHU-Hi-LongKou dataset (with 0.1\% training samples).} 
  \label{WH_vis} %
\end{figure*}

\begin{figure*}[!t] % 双栏图片
  \centering % 居中
  \includegraphics[width=1\linewidth,height=0.6\textheight]{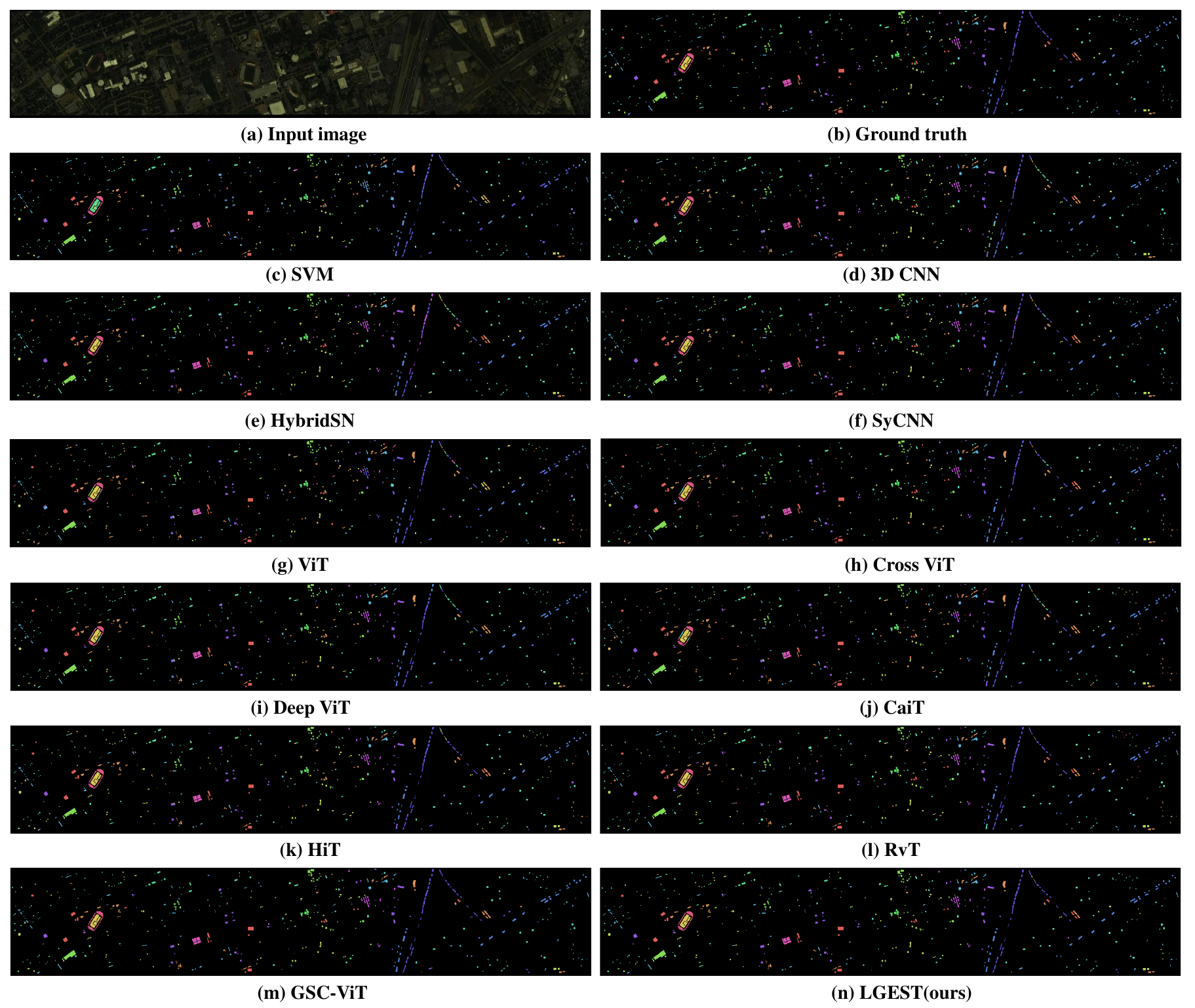}% 图片地址，可以pdf可以jpg，scale是缩放比例
  \caption{Classification map of different methods on the Houston2013 dataset (with 10\% training samples).} % 图片标题
  \label{class_hous} %
\end{figure*}

To further illustrate the effectiveness of our method, Figs. \ref{class_ip}, \ref{class_ksc}, \ref{WH_vis} and \ref{class_hous} present the visualization results for all classification methods. The figures illustrate that the classification maps produced by other methods exhibit more noise and greater confusion at the boundaries between different categories, whereas the classification maps generated by our proposed LGEST method are smoother and demonstrate more accurate classification results overall.

As shown in Fig. \ref{class_ip}, SVM produces the poorest results with frequent misclassifications. Convolution-based methods, limited to local features, misclassify boundary regions, while transformers introduce noise and struggle with classes of small coverage due to missing local details. Transformer–convolution hybrids improve accuracy but still fail on small classes. In contrast, LGEST produces refined maps that align closely with the ground truth.

% \cite{van2014accelerating}
\subsubsection{Analysis of t-SNE results}\label{sec4.2.3}
\begin{figure}[ht!] % 双栏图片
        \centering % 居中
        \includegraphics[width=0.5\textwidth, height=0.21\textheight]{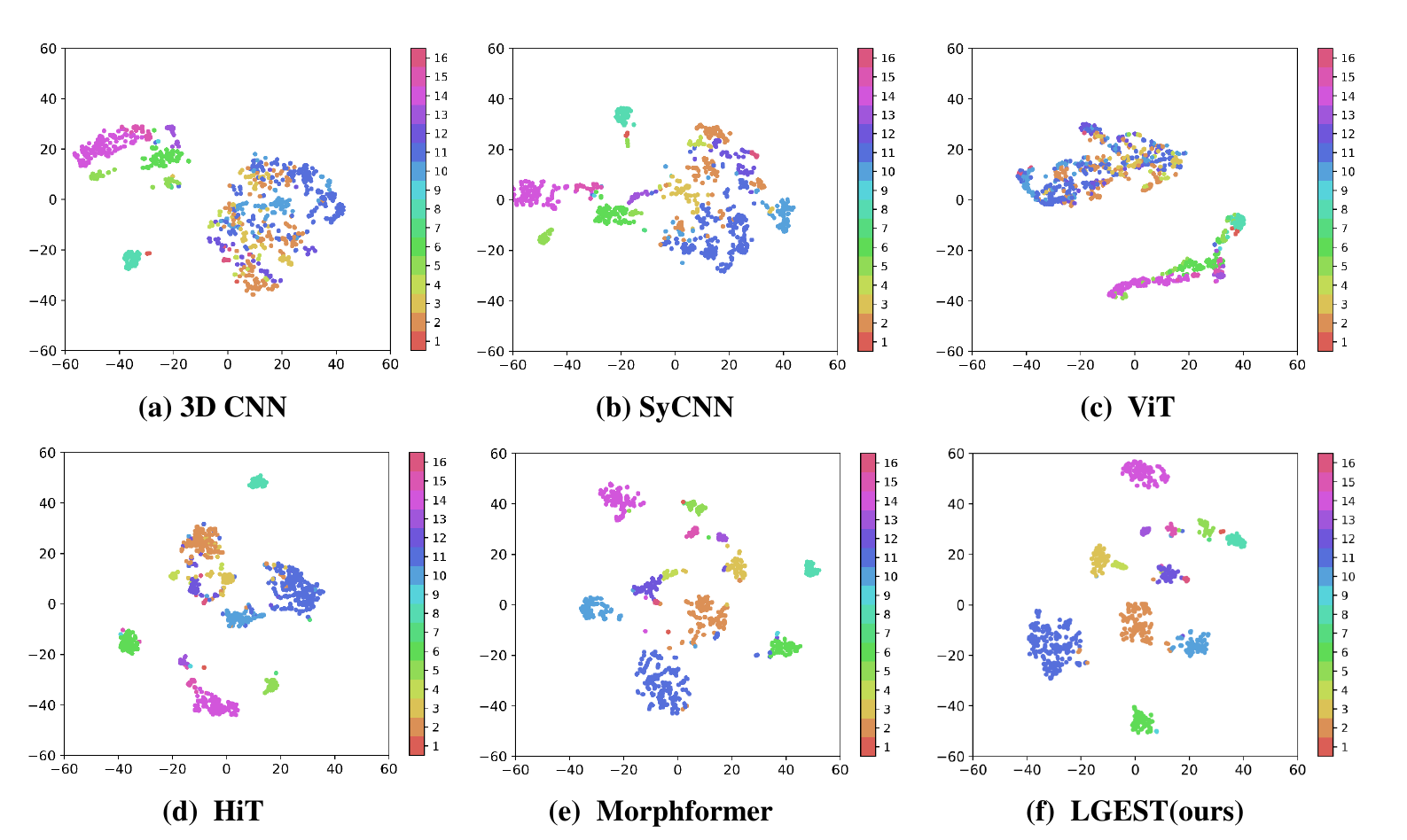}% 图片地址，可以pdf可以jpg，scale是缩放比例
        \caption{Visualization of t-SNE data analysis on the IndianPines dataset.} % 图片标题
        \label{fig_tsne} %
    \end{figure}

To more effectively demonstrate the performance of each classifier, we selected five of the most representative classification methods for comparison with the proposed LGEST using t-distributed stochastic neighbor embedding (t-SNE). As shown in Fig. \ref{fig_tsne}, on the IndianPines dataset, our proposed method exhibits a smaller intra-class distance and a larger inter-class distance compared to other models, which suffer from higher intra-class confusion. In particular, the 3D CNN and ViT models yielded the poorest clustering results. Conversely, the SyCNN and HiT models demonstrated enhanced classification performance, although they still exhibited notable confusion. Morphformer exhibits clustering results comparable to those of the LGEST, yet displays a greater dispersion of scattered points and more considerable inter-class distances. In general, the results demonstrate that LGEST exhibits the most effective clustering and that the efficacy of clustering is positively correlated with classification performance.

\begin{table}[htp]
  \centering
  \caption{Complexity analysis on the Indian Pines dataset.}
  \label{tab7}
  \resizebox{\linewidth}{!}{
      {\fontsize{5}{6}\selectfont
      \begin{tabular}{ccccc}
      \hline
      Methods & F(G) & P(M)  & OA(\%)$\uparrow$  \\ \hline
      RvT & 21.11 & 8.94  & 90.14 $\pm$ 0.90 \\
      CaiT & 16.79 & 163.66  & 75.76 $\pm$ 1.52 \\
      SyCNN & 194.11 & 45.81  & 88.14 $\pm$ 1.19 \\
      LGEST & 108.32 & 32.99  & \textbf{93.69 $\pm$ 0.44} \\ \hline
      \end{tabular}
      }
  }
  \end{table}

\subsection{Complexity Analysis}
\label{sec4.3}

This section evaluates the computational efficiency of LGEST compared to representative methods using Floating Point Operations (FLOPs), number of parameters (Params), and Overall Accuracy (OA), as summarized in Table \ref{tab7}.

\textbf{Superior Efficiency over 3D CNNs:} 
Traditional 3D deep learning methods often suffer from high computational costs. As shown in Table \ref{tab7}, compared to the representative spatial-spectral model SyCNN, LGEST reduces FLOPs by approximately \textbf{44\%} (194.11 G vs. 108.32 G) and parameters by \textbf{28\%} (45.81 M vs. 32.99 M). Despite this reduction in complexity, LGEST achieves a substantial accuracy improvement of \textbf{5.55\%} over SyCNN. This demonstrates that LGEST's sparse expert routing mechanism is significantly more efficient at capturing discriminative features than dense 3D convolutions.

\textbf{Effective Parameter Utilization:}
When compared to Transformer-based methods, LGEST exhibits a superior balance between model capacity and performance. While CaiT possesses a massive parameter count (163.66 M), it fails to achieve competitive accuracy (75.76\%), indicating severe redundancy. In contrast, LGEST utilizes a moderate parameter scale (32.99 M) to achieve state-of-the-art performance (\textbf{93.69\%}). Although the lightweight RvT incurs lower FLOPs, its limited capacity creates a performance bottleneck at roughly 90\% OA. LGEST invests necessary computational resources to model complex local-global interactions, successfully breaking this bottleneck and offering the optimal trade-off for high-precision HSI classification tasks.

% \subsection{Complexity Analysis}\label{sec4.3}
% This section analyzes method complexity based on four key metrics: FLOPS, number of parameters, training time, and testing time. As shown in Table \ref{tab7}, "F" and "P" represent "FLOPS" and "Parameters", respectively. LGEST performs best in terms of OA but requires more computational resources and longer training time. The parameter increase is due to multiple CIEM module interactions and multiple expert subspaces in CIEM-FPN, which also increase training time. Despite increasing sparsity during inference, this advantage is limited. Future work will explore lightweight transformer schemes and more efficient MoE mechanisms. CaiT conserves resources by freezing patch embeddings but has suboptimal classification outcomes. In contrast, RvT shows more balanced performance. The inclusion of multiple sub-experts in LGEST increases the parameter burden, but their lightweight computations allow LGEST to outperform SyCNN in terms of FLOPS and parameters.

\subsection{Ablation studies}\label{sec4.4}

\subsubsection{Analysis of LGEST} \label{sec4.4.1}
In this section, we discuss the key components of LGEST and evaluate the functionality and effectiveness of its various modules using OA, AA, and $\kappa$ metrics. We conducted detailed experiments on the Deep Spatial-Spectral Auto-Encoder (DSAE), the Cross-Interactive Mixed Expert Feature Pyramid (CIEM-FPN), and the Local-Global Expert System(LGES), and the experimental results are presented in Table \ref{tab8}.
\begin{table}[htp]
\centering
\caption{Ablation studies on key components of LGEST.}
\label{tab8}
\resizebox{\linewidth}{!}{
\begin{tabular}{cccc}
\hline
Methods & OA(\%)$\uparrow$ & AA(\%)$\uparrow$ & Kappa(\%)$\uparrow$ \\ \hline
DSAE & 91.50 & 74.49 & 90.30 \\
DSAE w/Local Expert Group & \textbf{93.05} & \textbf{77.25} & \textbf{92.05} \\ \hline\hline
\textbf{LGEST} & \textbf{93.67} & \textbf{79.54} & \textbf{92.76} \\
w/o CIEM-FPN & 92.82 & 72.54 & 91.78 \\
w/o Local Expert Group & 93.28 & 77.31 & 92.31 \\
w/o Global Expert Group & 93.33 & 78.60 & 92.37 \\
w/o Local-Global Expert System & 92.87 & 76.55 & 91.86 \\

\hline
\end{tabular}
}
\end{table}

\begin{figure*}[!htp] % 双栏图片
    \centering % 居中
    \includegraphics[width=1\textwidth]{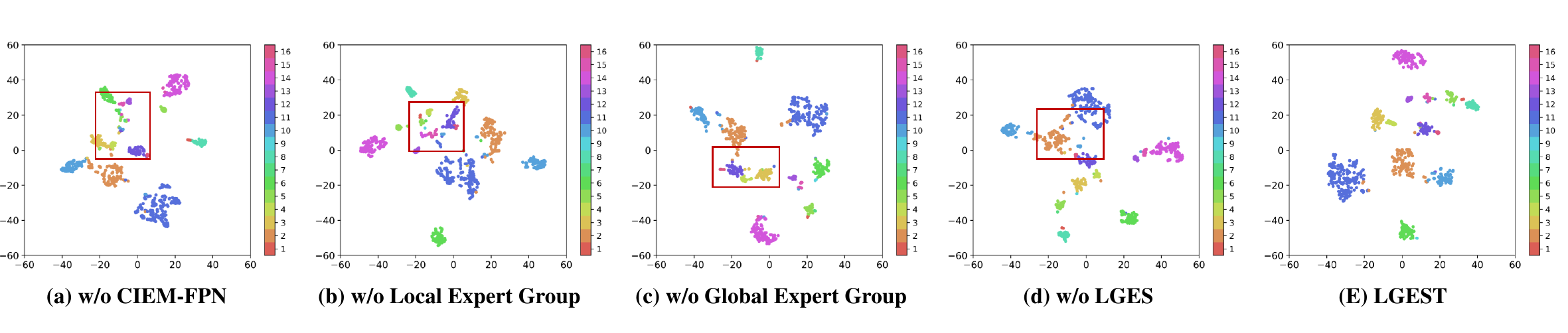}% 图片地址，可以pdf可以jpg，scale是缩放比例
    \caption{Visualization of t-SNE data analysis on different modules in LGEST. The area marked by the red box is the confusing part.} % 图片标题
    \label{fig_tsneab} %
\end{figure*}

Table \ref{tab8} shows the poorest model results were from DSAE alone due to its lack of global spatial-spectral extraction capability. However, after adding the Local Expert Group (LEG), the overall accuracy (OA) significantly improved. For the Local Global Expert System (LGEST), the inclusion of CIEM-FPN, with its dual-branch Cross-Interactive module for global spatial-spectral feature fusion, had a significant impact on the average accuracy (AA) value. The AA decreased after removing the local-global expert system. To explore the impact of the local and global expert groups, separate tests were conducted, with similar results, demonstrating their comparable importance. Visualized t-SNE experiments were conducted to demonstrate the effect of different components on classification performance. The results (as shown in Fig. \ref{fig_tsneab}) show that the complete LGEST model significantly outperforms the removal of any individual module in terms of feature clustering.

\subsubsection{Ablation study of the number of experts}\label{sec4.4.2}
% In this experiment, we investigated the effect of varying the number of experts across four datasets. The number of experts dictates the model's capacity to process complex data; however, merely increasing this number does not guarantee improved performance. When the number of experts becomes excessively large, the model's parameter count increases substantially, leading to expert competition. This phenomenon arises when multiple experts process similar data, resulting in insufficient training for each expert and challenges in gating selection due to data similarity. Conversely, if the number of experts is too small, each sub-expert must process highly diverse data types, thereby degrading model performance.

% As shown in Table \ref{number_of_experts}, the optimal number of experts for the model varies depending on the dataset. The number column represents the number of local and global experts. Specifically, the optimal number of experts on the Indian Pines and Houston2013 datasets is 4, while the optimal number on KSC is 2 and WHU-Hi-LongKou is 8. Overall, the model demonstrates more balanced performance on mainstream HSI datasets when 4 experts are used.
This experiment explored the effect of changing the number of experts on four datasets. The number of experts affects the model's ability to process complex data, but adding more experts doesn't always improve performance. If there are too many experts, the model has a lot of parameters, which can cause expert competition. This happens when multiple experts work on similar data, which can result in each expert not getting enough training. On the other hand, if there are too few experts, each expert has to work on very different types of data, which can also hurt performance. Table \ref{number_of_experts} shows that the optimal number of experts varies for each dataset. Overall, the model performs best when there are four experts.

\begin{table}[h]
\renewcommand{\arraystretch}{1.2}
    \centering
    \caption{Ablation study on the number of experts. (best results in bold)} \label{number_of_experts}
    \resizebox{\linewidth}{!}{
    \begin{tabular}{ccccccc}
    \hline
    Datasets & Number & F(G) & P(M) & OA(\%)$\uparrow$ & AA(\%)$\uparrow$ & Kappa(\%)$\uparrow$ \\ \hline
    \multirow{4}{*}{Indian Pines}
    & [2,2] & 107.26 & 22.36 & 93.41$\pm$0.61 & 80.48$\pm$2.41 & 92.47$\pm$0.70 \\
    & \textbf{[4,4]} & 108.32 & 32.99 & \textbf{93.69$\pm$0.44} & 80.72$\pm$3.00 & \textbf{92.80$\pm$0.50} \\
    & [8,8] & 110.45 & 54.24 & 93.56$\pm$1.11 & \textbf{80.78$\pm$4.13} & 92.63$\pm$0.45 \\
    & [16,16] & 114.70 & 96.75 & 93.45$\pm$0.47 & 79.53$\pm$3.49 &92.51$\pm$0.54 \\ \hline

    \multirow{4}{*}{KSC}
    & \textbf{[2,2]} & 106.77 & 20.58 & \textbf{92.59$\pm$1.44} & \textbf{90.32$\pm$1.89} & \textbf{91.75$\pm$1.60} \\
    & [4,4] & 107.66 & 29.43 & 91.44$\pm$1.21 & 88.63$\pm$1.66 & 90.46$\pm$1.34 \\
    & [8,8] & 109.43 & 47.14 & 91.24$\pm$2.12 & 88.05$\pm$3.22 & 90.24$\pm$2.36 \\
    & [16,16] & 112.97 & 82.56 & 90.86$\pm$1.25 & 87.28$\pm$1.91 & 89.82$\pm$1.40 \\ \hline

    \multirow{4}{*}{Houston2013}
    & [2,2] & 106.47 & 21.74 & 95.58$\pm$0.44 & 95.67$\pm$0.52 & 95.22$\pm$0.48 \\
    & \textbf{[4,4]} & 107.48 & 31.78 & \textbf{95.86$\pm$0.58} & \textbf{95.88$\pm$0.40} & \textbf{95.53$\pm$0.63} \\
    & [8,8] & 109.49 & 51.85 & 95.51$\pm$0.41 & 95.43$\pm$0.53 & 95.14$\pm$0.44 \\
    & [16,16] & 113.50 & 92.00 & 95.77$\pm$0.49 & 95.67$\pm$0.62 & 95.42$\pm$0.53 \\ \hline

    \multirow{4}{*}{WHU-Hi-LongKou}
    & [2,2] & 107.75 & 18.27 & 94.55$\pm$1.02 & 79.16$\pm$5.78 & 92.78$\pm$1.37 \\
    & [4,4] & 108.40 & 24.76 & 95.17$\pm$0.87 & 81.55$\pm$3.84 & 93.61$\pm$1.16 \\
    & \textbf{[8,8]} & 109.70 & 37.75 & \textbf{95.40$\pm$0.50} & \textbf{82.14$\pm$3.19} & \textbf{93.91$\pm$0.66} \\
    & [16,16] & 112.30 & 63.73 & 94.44$\pm$1.10 & 77.75$\pm$4.70 & 92.62$\pm$1.50 \\
    \hline
    \end{tabular}
    }
\end{table}

\begin{figure}[!htp] % htbp详见说明书，记得删除括号内容
  \centering % 居中
  \includegraphics[width=0.5\textwidth]{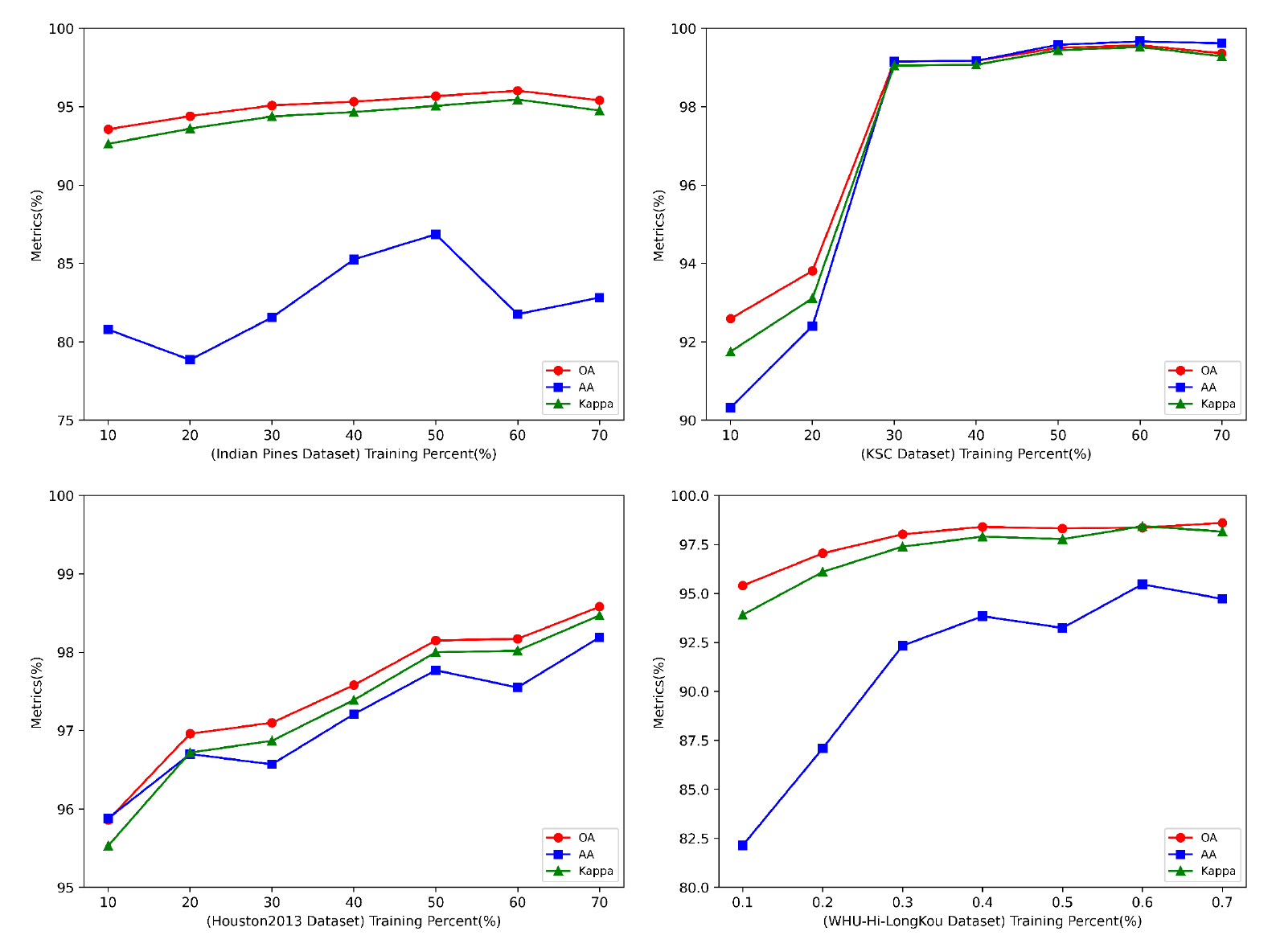}% 图片地址，可以pdf可以jpg，scale是缩放比例
  \caption{Ablation study with training percentages on four datasets.} % 图片标题
  \label{fig_tsamp} %
\end{figure}

\subsubsection{Ablation study of image patch size} \label{sec4.4.3}
  % The patch size in the HSI classification task is the image block size of each input data. To verify the performance of LGEST under different patch sizes, we conducted experiments on four standard datasets with different patch sizes. As shown in Table \ref{tab9}, the optimal patch size of LGEST differs across datasets. With the number of experts fixed at 8, the optimal patch size for Indian Pines is 11 $\times$  11, for KSC is 17 $\times$  17, for Houston2013 is 9 $\times$  9, and for WHU-Hi-LongKou is 15 $\times$ 15. Notably, at a patch size of 15 $\times$  15, the model demonstrates more consistent performance across datasets.
The patch size in LGEST critically influences the balance between local detail preservation and global context capture, with optimal sizes varying according to the spatial characteristics of each dataset and the hierarchical architecture of the model. As shown in Table \ref{tab9}, this variation demonstrates how the model adapts to different spatial configurations through its dual architecture: DSAE effectively compresses input patches while multi-scale fusion in CIEM-FPN and expert routing in LGES dynamically balance local and global information. This ablation study validates the robustness of LGEST across varying patch sizes, demonstrating its adaptive capacity to handle diverse spatial configurations through its hierarchical expert system.

\begin{table}[htp]
\renewcommand{\arraystretch}{1.15}
    \centering
    \caption{Ablation study on the patch sizes (best results in bold)}\label{tab9}
    \resizebox{\linewidth}{!}{
    \begin{tabular}{ccccc}
    \hline
    Datasets & Patch Size & OA(\%)$\uparrow$ & AA(\%)$\uparrow$ & Kappa(\%)$\uparrow$ \\ \hline
    \multirow{5}{*}{Indian Pines}
    & 9×9 & 95.07$\pm$0.74 & \textbf{86.86$\pm$0.77} & 94.37$\pm$0.85 \\
    & \textbf{11×11} & \textbf{95.50$\pm$0.32} & 83.07$\pm$0.96 & \textbf{94.87$\pm$0.36} \\
    & 13×13 & 95.01$\pm$0.65 & 84.55$\pm$2.80 & 94.30$\pm$0.74 \\
    & 15×15 & 93.56$\pm$1.11 & 80.78$\pm$4.13 & 92.63$\pm$0.45 \\
    & 17×17 & 91.69$\pm$0.72 & 75.48$\pm$1.44 & 90.51$\pm$0.82 \\ \hline

    \multirow{5}{*}{KSC}
    & 9×9 & 89.21$\pm$1.21 & 83.28$\pm$1.72 & 87.97$\pm$1.35 \\
    & 11×11 & 90.27$\pm$0.96 & 84.10$\pm$2.15 & 89.14$\pm$1.07 \\
    & 13×13 & 93.34$\pm$0.53 & 90.09$\pm$0.49 & 92.58$\pm$0.60 \\
    & 15×15 & 91.24$\pm$2.12 & 88.05$\pm$3.22 & 90.24$\pm$2.36 \\
    & \textbf{17×17} & \textbf{93.36$\pm$1.33} & \textbf{91.17$\pm$2.79} & \textbf{92.60$\pm$1.49} \\ \hline

    \multirow{5}{*}{Houston2013}
    & \textbf{9×9} & \textbf{97.77$\pm$0.21} & \textbf{97.59$\pm$0.34} & \textbf{97.59$\pm$0.23} \\
    & 11×11 & 97.56$\pm$0.40 & 97.38$\pm$0.59 & 97.36$\pm$0.43 \\
    & 13×13 & 97.15$\pm$0.22 & 96.78$\pm$0.46 & 96.92$\pm$0.24 \\
    & 15×15 & 95.51$\pm$0.41 & 95.43$\pm$0.53 & 95.14$\pm$0.44 \\
    & 17×17 & 94.84$\pm$0.51 & 94.79$\pm$0.61 & 94.43$\pm$0.55 \\ \hline

    \multirow{5}{*}{WHU-Hi-LongKou}
    & 9×9 & 93.84$\pm$0.80 & 74.64$\pm$5.13 & 91.81$\pm$1.08 \\
    & 11×11 & 93.29$\pm$0.76 & 72.84$\pm$2.86 & 91.09$\pm$0.99 \\
    & 13×13 & 94.74$\pm$0.52 & 78.79$\pm$2.00 & 93.05$\pm$0.70 \\
    & \textbf{15×15} & \textbf{95.40$\pm$0.50} & \textbf{82.14$\pm$3.19} & \textbf{93.91$\pm$0.66} \\
    & 17×17 & 95.06$\pm$0.67 & 79.40$\pm$3.62 & 93.44$\pm$0.91 \\
    \hline
    \end{tabular}
    }
\end{table}

\subsubsection{Ablation study of the percentage of training samples} \label{sec4.4.4}
In this section, we conduct ablation experiments to analyze the effect of training sample percentages on the four datasets. The experimental results of the proposed LGEST are shown in Fig. \ref{fig_tsamp}. The results indicate that the overall OA value increases as the number of training samples rises. This trend is particularly prominent in the KSC dataset, where the OA value approaches 100\% when the training samples reach 50\%. Notably, the AA metrics for the Indian Pines dataset show a sharp decline as the training samples increase to 60\%. We speculate that the increase in larger class samples negatively affects the classification accuracy of smaller classes.  Overall, increasing the amount of data enables LGEST to capture more hyperspectral features.

\section{Conclusion}\label{conclusion}
This paper introduces the Local-Global Expert Spatial-Spectral Transformer (LGEST), a novel framework for hyperspectral image (HSI) classification that overcomes key limitations of existing approaches through three pivotal innovations. First, the Deep Spatial-Spectral Autoencoder (DSAE) hierarchically compresses three-dimensional HSI cubes into compact, discriminative embeddings while preserving neighborhood coherence, thereby mitigating information loss in high-dimensional spaces. Second, the Cross-Interactive Mixed Expert Feature Pyramid (CIEM-FPN) employs cross-attention mechanisms and residual mixture-of-experts layers to dynamically fuse multi-scale features by adaptively weighting spectral and spatial cues via learnable gating. Third, the Local-Global Expert System (LGES) utilizes sparsely activated expert pairs—where convolutional sub-experts capture fine-grained textures and transformer sub-experts model long-range dependencies—combined with confidence-aware routing for adaptive processing. Extensive experiments on four benchmark datasets demonstrate that LGEST achieves superior overall accuracy, average accuracy, and $\kappa$ coefficient compared to state-of-the-art CNN and Transformer methods.

Future work will address the computational complexity of LGEST by investigating lightweight Transformer architectures and more efficient mixture-of-experts mechanisms, such as dynamic expert pruning and parameter sharing strategies. Finally, exploring few-shot learning scenarios with LGEST remains a priority, leveraging its expert-based architecture to improve generalization with limited labeled samples and thereby broadening its applicability to real-world remote sensing applications.

\section*{Data availability}
Data will be made available on request.

\section*{Declaration of competing interest}
The authors declare that they have no known competing financial interests or personal relationships that could have appeared to influence the work reported in this paper.

\section*{Funding}
This work was jointly supported by the following projects: National Natural Science Foundation of China (NSFC) Fund under Grant 62301174 and 62403155. National Key Research and Development Program of China Fund under Grant 2021YFB3901204 and 2024YFC3015605. Guangzhou basic and applied basic research topics under Grant 2024A04J2081 and 2024A04J3369.

\bibliographystyle{elsarticle-num}
\bibliography{main}
\end{document}